\documentclass[lettersize,journal]{IEEEtran}
\usepackage{amsmath,amsfonts}
\usepackage{algorithmic}
\usepackage{algorithm}
\usepackage{array}
\usepackage[caption=false,font=normalsize,labelfont=sf,textfont=sf]{subfig}
\usepackage{textcomp}
\usepackage{stfloats}
\usepackage{url}
\usepackage{verbatim}
\usepackage{graphicx}
\usepackage{cite}
\usepackage{algorithm}
\usepackage{algorithmic}
\usepackage{tabularx}
\usepackage{multirow}
\usepackage{graphicx}
\usepackage{booktabs}
\usepackage{booktabs, multirow}
\usepackage[most]{tcolorbox}
\usepackage{tabularx}
\usepackage{enumitem}
\usepackage{enumitem}
\usepackage{amsmath}
\usepackage{amssymb}
\usepackage{array}
\usepackage{graphicx}

\begin{document}

\title{MemeLogos: Knowledge-Driven Retrieval-Augmented Reasoning for Harmful Meme Detection}

\author{
Hexiang~Gu,
Saihui~Hou,~\IEEEmembership{Member,~IEEE},
Changlan~Liu,
Min~Ren,
Huijia~Wu,
Hongxun~Gu,
and~Zhaofeng~He,~\IEEEmembership{Member,~IEEE}%

\thanks{Hexiang Gu, Min Ren, Huijia Wu, and Zhaofeng He are with
Beijing University of Posts and Telecommunications, Beijing 100876, China
(E-mail: guhexiang@bupt.edu.cn; minren@bupt.edu.cn; huijiawu@bupt.edu.cn;
 zhaofenghe@bupt.edu.cn).}
\thanks{
Saihui Hou is with the School of Artificial Intelligence, Beijing Normal University, Beijing 100875, China. (Email: housaihui@bnu.edu.cn)
}
\thanks{Changlan Liu is with the College of Electrical Engineering, Zhejiang University, Zhejiang 310027, China, and also with Zhongguancun Academy, Beijing, China. E-mail: liuchanglan@zju.edu.cn}
\thanks{Hongxun Gu is with China Telecommunications Corporation, Beijing 100033, China. E-mail: guhongxun@163.com}

}
%

\markboth{IEEE Transactions on Knowledge and Data Engineering}%
{Hexiang Gu \MakeLowercase{\textit{et al.}}: MemeLogos: Knowledge-Driven Retrieval-Augmented Reasoning for Harmful Meme Detection}


\maketitle

\begin{abstract}
With the rapid proliferation of social media, harmful memes have become a growing threat to online ecosystems. Their harmfulness is often implicitly conveyed through complex interactions among visual, textual, and contextual cues, making accurate detection challenging. Recent studies have introduced Retrieval-Augmented Generation (RAG) and external knowledge into harmful meme detection. However, existing methods mainly retrieve similar instances or auxiliary information, limiting their ability to capture reusable harmfulness knowledge and underlying reasoning mechanisms, and thus constraining both detection performance and interpretability. To address these challenges, we propose MemeLogos, a knowledge-driven retrieval-augmented reasoning framework for harmful meme detection. MemeLogos abstracts recurring harmfulness patterns and reasoning structures from diverse harmful cases into reusable knowledge units and organizes them into a dedicated star-shaped knowledge graph. By combining structured knowledge retrieval with an explicit multi-stage reasoning process inspired by cognitive mechanisms, MemeLogos leverages retrieved knowledge to guide harmfulness reasoning, improving both detection reliability and interpretability.
Extensive experiments show that MemeLogos consistently outperforms existing harmful meme detection methods while generating more coherent and interpretable reasoning, demonstrating the effectiveness of structured harmfulness knowledge for multimodal content understanding.
\end{abstract}


\noindent
{\small\textcolor{red}{\textit{Warning: This paper contains potentially disturbing meme examples involving harmful content, presented solely for research purposes.}}}

\section{Introduction}
\IEEEPARstart{U}{nlike} unimodal harmful content, memes often convey implicit harmful meanings through subtle interactions between visual and textual cues, such as vulgarity, sarcasm, and violence~\cite{Kiela2020TheHM}. For example, juxtaposing a victim's photo with historical tragedy symbols lacks surface-level offensive words, yet infuses malicious historical irony. Since harmful semantics often depend on contextual knowledge, cross-modal compositionality and implicit logical relations, black-box multimodal alignment alone is insufficient for reliable judgement~\cite{ijcai2024p684}. Therefore, effective harmful meme detection requires not only cross-modal semantic understanding, but also the ability to organize contextual evidence into reusable harmfulness knowledge and reason over their structured semantic relations~\cite{garg-etal-2025-just}.

Harmful meme detection relies heavily on the mining of contextual knowledge and harmful judgement logic~\cite{Li2026KIDKD, Pan2025ReadAY}. Although current solutions have adopted retrieval augmentation, explanation and reasoning-driven detection, they still suffer from two core limitations: (1) \textbf{Insufficiency of performance.} Existing methods rely on similar case retrieval and explanatory insights from datasets, with knowledge organized at the instance or free-form text level, which rarely distill recurring underlying harmful patterns, making it hard to model stable logical relations across diverse meme scenarios, thereby restricting further performance improvement of the model. (2) \textbf{Limited quality of interpretability.} Existing explanation-oriented methods often justify predictions with retrieved cases or free-form rationales. However, such evidence is mostly instance-level and rarely reveals the essential harmfulness rule triggered by the meme, limiting a transparent reasoning path from multimodal cues to verdicts and weakening the traceability and auditability of interpretation.

\begin{figure*}[t]  
  \centering
  \includegraphics[width=\textwidth]{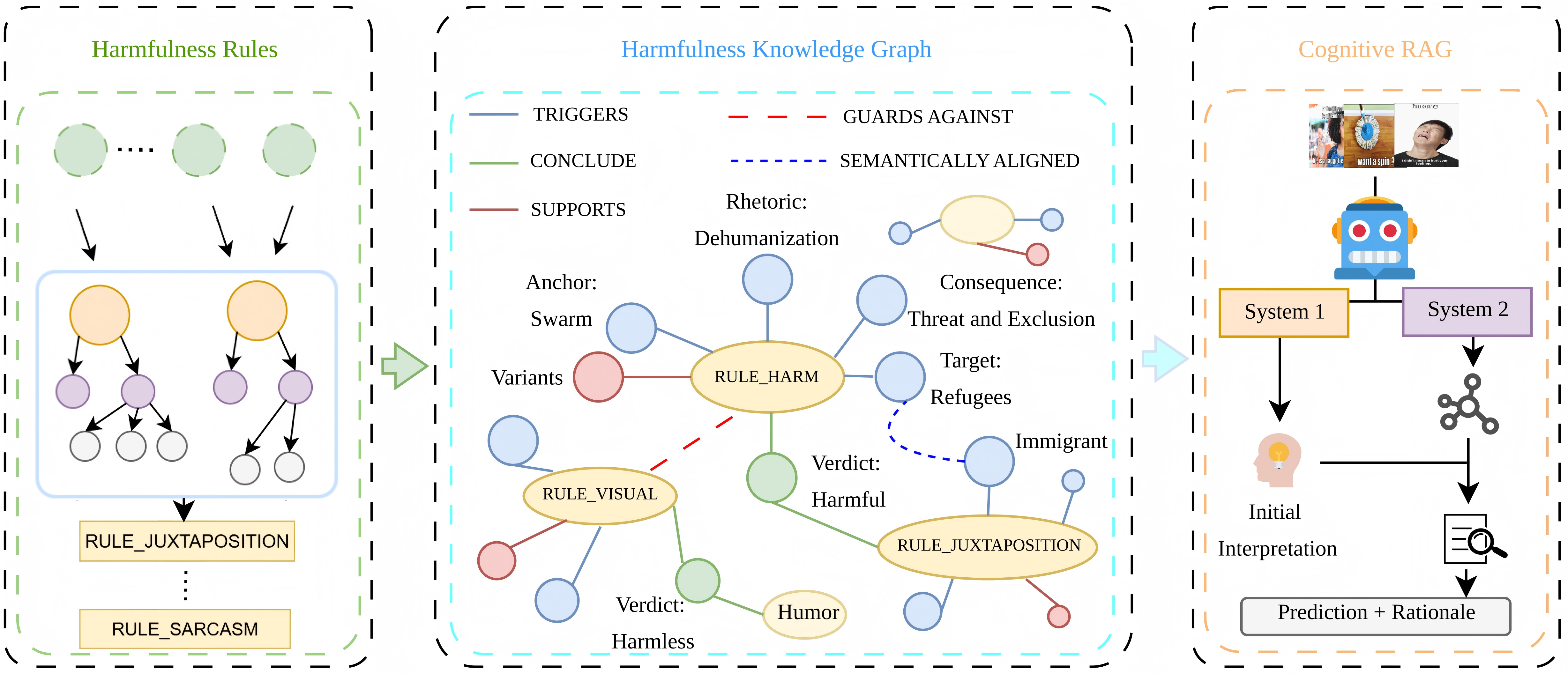}
  \caption{Overview of MemeLogos. The framework abstracts reusable harmfulness rules from multimodal meme cases, organizes them into a structured harmfulness knowledge graph, and performs graph-grounded retrieval and cognitive reasoning for accurate and interpretable harmful meme detection.}
  \label{fig:method}
\end{figure*}

Driven by the above observations, we propose MemeLogos, a reasoning framework for harmful meme detection, incorporating structured harmfulness knowledge with Retrieval-Augmented Generation (RAG) reasoning paradigm to establish interpretable and effective harmful judgement logics. 
We distill reusable rule units from harmful meme samples with Chain-of-Thought (CoT) annotations,
to derive structured harmfulness knowledge, which is further organized into a harmfulness knowledge graph with a star-shaped topology. This graph-organized harmfulness knowledge provides two key advantages: (1) \textbf{Comprehensive and Reusable Harmfulness Rules.} The extracted rule units abstract recurring harmfulness patterns and underlying semantic mechanisms from diverse multimodal memes, capturing generalizable harmfulness knowledge beyond instance-specific visual-textual expressions. By preserving the essential semantic associations and judgment criteria shared across different scenarios, these rules reduce reliance on individual instances and support more consistent and reusable harmfulness reasoning. (2) \textbf{Fine-grained Logical Association.} By explicitly organizing the semantic attributes, supporting evidence, and verdicts of harmfulness rules, the knowledge graph establishes traceable associations between harmfulness patterns and their judgments. Furthermore, cross-rule semantic relations facilitate knowledge expansion and contrastive retrieval between related rules with opposing verdicts, providing structured support for distinguishing subtle harmfulness boundaries and enabling more fine-grained harmfulness reasoning.

To fully leverage graph-organized harmfulness knowledge, we draw inspiration from Dual-System Theory of cognitive science~\cite{kahneman2011thinking}, which posits that human cognition is driven by a fast, intuitive system and a slow, deliberative system. Building on this perspective, we propose Cognitive RAG, which dynamically retrieves relevant rules from the harmfulness knowledge graph and performs contrastive reasoning. The framework mitigates the risk of conventional RAG being misled by noisy or only superficially relevant retrieved evidence~\cite{yu2024chain, amiraz2025distracting}, which exhibits two distinct core advantages: (1) \textbf{Superior performance.} Such logical modeling significantly enhances generalizability across diverse domains and improves the overall performance of harmful meme detection. (2) \textbf{Deep interpretability.} The reasoning chain generated by MemeLogos transforms complex metaphor recognition into a traceable and structured reasoning process, thereby yielding a substantial improvement in interpretability.

The main contributions of this paper can be summarized as follows:

\begin{itemize}
\item We construct structured harmfulness knowledge and organize it as a knowledge graph, inducing reusable rule-level knowledge from richly annotated harmful meme data and providing a structured basis for decomposing the underlying composition of harmful meaning.
\item We propose MemeLogos, a RAG framework that formalizes harmful meme detection through knowledge reasoning. By bridging structured reasoning with RAG, MemeLogos significantly enhances detection performance while providing traceable and interpretable reasoning.
\item Extensive experiments on harmful meme dataset show that MemeLogos consistently improves performance across different backbone models and outperforms existing state-of-the-art methods, achieving 86.96\% accuracy on MemeMind. With training alignment, MemeLogos\_Align achieves 89.50\% accuracy, demonstrating the effectiveness of our framework.

\end{itemize}

\section{Related Work}

\subsection{Harmful Meme Detection}



Harmful meme detection has become an important research topic for online content safety. Benchmark datasets such as Hateful Memes~\cite{Kiela2020TheHM}, HarMeme~\cite{pramanick-etal-2021-detecting}, MAMI~\cite{fersini-etal-2022-semeval}, and ToxiCN\_MM~\cite{Lu2024TowardsCD} have established diverse evaluation settings covering different harmfulness categories, languages, and social contexts. Early approaches primarily focused on multimodal feature extraction and fusion, while subsequent dual-stream architectures and vision-language pre-training models~\cite{Dosovitskiy2020AnII,Radford2021LearningTV} enabled more effective cross-modal representation learning. For instance, Hate-CLIPper~\cite{Kumar2022HateCLIPperMH} explicitly models interactions between CLIP-based visual and textual features to capture their complementary information for harmfulness classification.

Nevertheless, harmful meanings are frequently conveyed through implicit cross-modal interactions involving irony, metaphor, cultural references, and contextual incongruity, making them difficult to recognize through surface-level correlations alone. To address these challenges, recent studies have incorporated auxiliary semantic understanding and reasoning into harmful meme detection. CapAlign~\cite{Ji2024CapAlignIC} employs question-driven informative caption generation to enrich visual semantics and improve cross-modal alignment. Other approaches leverage contextualized captioning and chain-of-thought reasoning to uncover implicit meanings and support interpretable predictions~\cite{10.1145/3701716.3718385,Pan2025ReadAY}. Furthermore, Lin et al.~\cite{lin-etal-2023-beneath} introduce multimodal reasoning distillation to transfer implicit semantic reasoning from large language models into harmful meme detectors. More recently, ExplainHM++~\cite{lin2025explainhm++} integrates similar meme retrieval with multimodal debate to enhance harmfulness detection and interpretability through competing harmful and harmless interpretations.

Despite these advances, much of the existing reasoning and retrieval process remains centered on individual memes, free-form explanations, or retrieved examples. Although such information can enrich contextual understanding, recurring harmfulness patterns and their relationships across different scenarios are not always explicitly abstracted into reusable rule-level knowledge. Consequently, connecting multimodal semantic cues with generalizable harmfulness criteria and supporting evidence remains challenging, motivating more structured knowledge representation and reasoning for harmful meme detection.

\subsection{Graph-Based Retrieval-Augmented Generation}
Recent studies extend conventional Retrieval-Augmented Generation (RAG) by explicitly modeling structured relations among external knowledge units. GraphRAG~\cite{edge2024graphrag} constructs entity-centric graph indexes and community summaries to support global reasoning over large document collections, while G-Retriever~\cite{he2024gretriever} formulates retrieval over textual graphs as a subgraph selection problem. HippoRAG~\cite{gutierrez2024hipporag} combines knowledge graphs with graph-based propagation to support multi-hop knowledge integration, and KG$^2$RAG~\cite{zhu2025kg2rag} exploits knowledge-graph relations to expand and organize semantically retrieved evidence. Beyond retrieval, Think-on-Graph~\cite{sun2024tog} and Reasoning on Graphs~\cite{luo2024rog} explore explicit graph traversal and relation-path reasoning over knowledge graphs, while Think-on-Graph 2.0~\cite{ma2025tog2} further integrates structured graph retrieval with unstructured contextual evidence. StructRAG~\cite{li2025structrag} additionally demonstrates the benefit of dynamically organizing retrieved knowledge into task-appropriate structured representations. These studies highlight the importance of relational structure for knowledge retrieval and reasoning. Different from general Graph-RAG methods that mainly operate over entity-, document-, or fact-level structures, MemeLogos abstracts multimodal harmful cases into reusable rule-level knowledge and explicitly models semantic and polarity-aware relations among harmfulness rules, enabling task-specific retrieval and contrastive reasoning for harmful meme detection.

\begin{table*}[t]
\caption{Statistics of the harmfulness knowledge graph.}
\label{tab:graph_statistics}
\centering
\small
\setlength{\tabcolsep}{4.6pt}
\renewcommand{\arraystretch}{1.05}

\begin{tabular}{@{}lrrrrrrrr@{\hspace{4pt}}lrrrrrr@{}}
\toprule
\multicolumn{9}{c}{\textbf{Nodes}}
&
\multicolumn{7}{c}{\textbf{Edges}}\\
\cmidrule(r){1-9}
\cmidrule(l){10-16}

\textbf{Type}
&\textbf{Case}
&\textbf{Rule}
&\textbf{Anch.}
&\textbf{Rhet.}
&\textbf{Targ.}
&\textbf{Cons.}
&\textbf{Verd.}
&\textbf{Total}
&
\textbf{Relation}
&\textbf{Align.}
&\textbf{Trig.}
&\textbf{Guard}
&\textbf{Supp.}
&\textbf{Concl.}
&\textbf{Total}
\\

\midrule

\textbf{Count}
&4477
&2541
&2195
&2056
&2002
&1829
&2541
&17641
&
\textbf{Count}
&15706
&10164
&8084
&4477
&2541
&40972
\\

\bottomrule
\end{tabular}

\end{table*}

\section{Methodology}

To boost the performance and interpretability of harmful meme detection, we propose MemeLogos, a reasoning-oriented detection framework, which integrates graph-based structured harmfulness knowledge and retrieval-augmented reasoning technology. The overall architecture is illustrated in Figure~\ref{fig:method}. The framework consists of two core stages: (1) \textbf{Harmfulness Knowledge} construction stage, we extract inherent harmfulness logic rules and build the knowledge graph for harmfulness detection; (2) \textbf{Graph-Grounded Cognitive RAG} inference stage, we combine adversarial graph retrieval and dual-system audit reasoning mechanism to achieve superior performance and auditable reasoning.

\begin{figure}[t]
  \centering
  \includegraphics[width=\columnwidth]{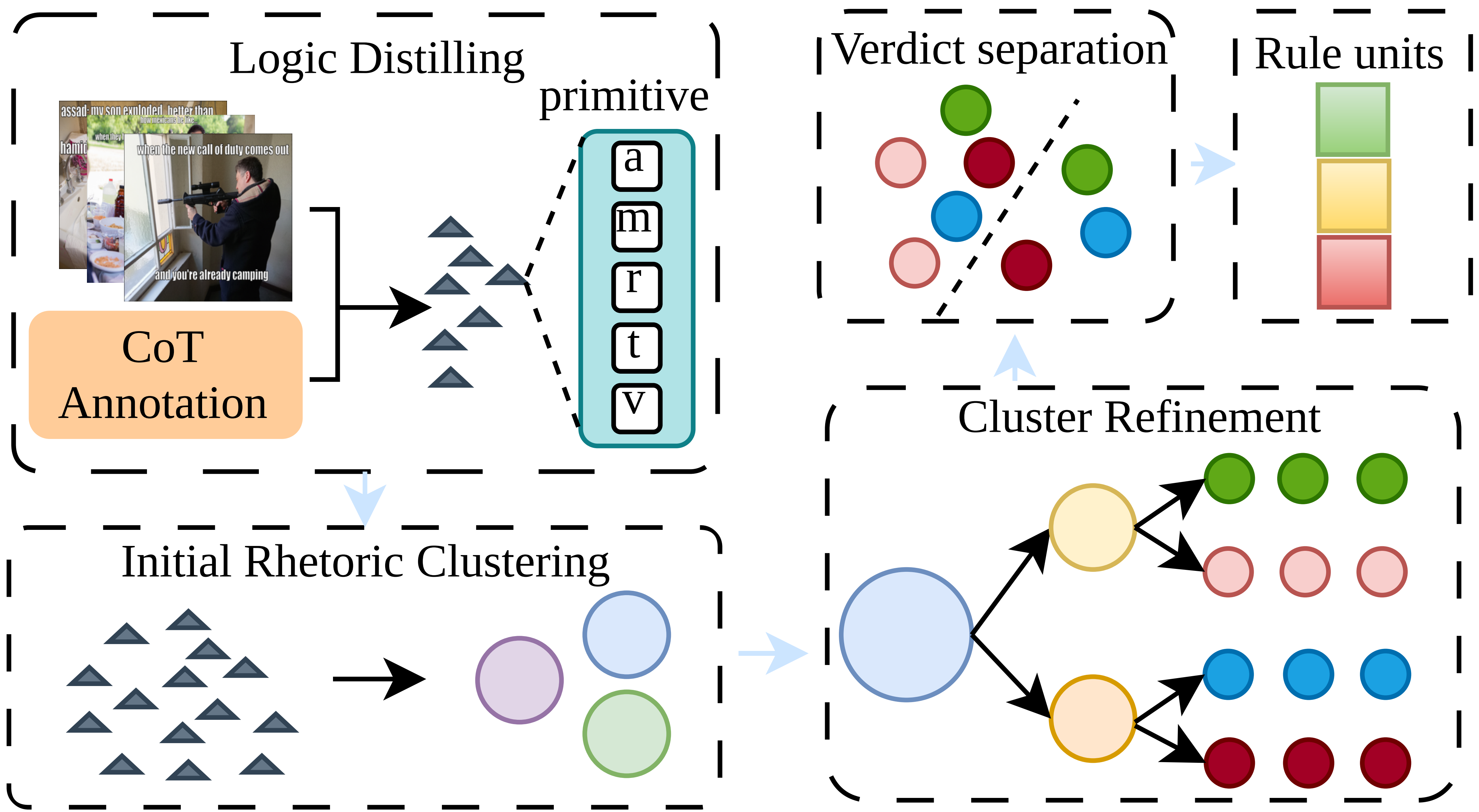}
  \caption{Pipeline of Formalized Rule Extraction. MemeLogos first distills each meme into structured harmfulness primitives, then performs rhetoric-based clustering, semantic cluster refinement and verdict-aware separation to aggregate recurring patterns, which are finally generalized into reusable harmfulness rule units.}
  \label{fig:rules}
\end{figure}

\subsection{Harmfulness Knowledge}

Existing methods often represent knowledge at the level of instances or unstructured textual explanations, making it difficult to capture inherent harmfulness patterns. Moreover, directly relying on models to generate knowledge from memes may introduce inherent semantic biases. To mitigate these issues, we ground harmfulness knowledge construction in meme instances and their detailed Chain-of-Thought (CoT) annotations. Specifically, leveraging the training data with detailed CoT annotations provided by MemeMind~\cite{Gu2025MemeMindAL}, we construct a harmfulness knowledge graph in two steps: 1) Formalized Rule Extraction, abstracting essential harmfulness patterns and rules from the memes with CoT annotations, and 2) Topological Graph Modeling, explicitly modeling harmfulness rules via star-shaped topology.

\subsubsection{Formalized Rule Extraction}

In this section, we leverage instance-level logic distillation and corpus-level rule generalization to map unstructured multimodal memes to formalized harmfulness rule units, as illustrated in Figure~\ref{fig:rules}, so as to accurately extract and generalize the inherent harmful logic hidden within heterogeneous memes.

\textbf{Instance-level logic distillation.}
Given a sample, we decompose the instance into six logical primitives:
\begin{equation}
z_i=(a_i,m_i,r_i,t_i,c_i,v_i),
\label{eq:logic_distilling}
\end{equation}
where $a_i$, $m_i$, $r_i$, $t_i$, $c_i$ and $v_i$  respectively denote the anchor, capturing the observable visual or textual carrier of the meme; semantic mapping, represents the latent semantic interpretation assigned to the carrier; rhetorical mechanism,  describing the compositional mechanism that connects the observable cue to the latent meaning; target, identifing the individual, group, institution, or social norm affected by the meme; consequence, recording the semantic or social effect and verdict, denoting the final harmfulness label. This logical decomposition reduces instance-specific multimodal noise while retaining the reasoning structure required for harmfulness assessment.

\textbf{Corpus-level rule generalization.}
Based on the Instance-Level Logic Distillation, we progressively cluster and canonicalize the extracted primitives to obtain generalizable rule units. As rhetorical mechanisms provide an important semantic basis for harmfulness patterns, 
we first use it to construct coarse semantic communities. For each
instance, the rhetorical primitive $r_i$ is encoded by MPNet~\cite{Song2020MPNetMA} as
\begin{equation}
\mathbf{h}^{r}_i=f_{\mathrm{MPNet}}(r_i),
\end{equation}
where $f_{\mathrm{MPNet}}$ denotes the MPNet semantic encoder.
Then we apply initial clustering on $\mathbf{h}^{r}_i$ based on cosine similarity with a threshold of $\tau_r$ . This produces coarse clusters whose members share similar rhetorical mechanisms and establishes a foundation for fine-grained rule induction.

However, clusters formed solely according to rhetorical mechanisms may remain
semantically heterogeneous, since the same rhetorical strategy can be instantiated
with different visual anchors, semantic mappings, targets, and consequences. We
therefore further refine each coarse cluster by incorporating these complementary
primitives. For each instance, the anchor, mapping, target, and consequence texts
are concatenated and encoded into a unified semantic representation $\mathbf h_i$:
\begin{equation}
\mathbf h_i=f_{\mathrm{MPNet}}
\left(a_i\oplus m_i\oplus t_i\oplus c_i\right),
\end{equation}
where $\oplus$ denotes text-level concatenation. Because coarse clusters vary substantially in size, the refinement
granularity is determined independently for each cluster. Let $\mathcal G_j$
denote the $j$-th coarse cluster and $n_j=|\mathcal G_j|$ its size. 
To further resolve semantic heterogeneity within coarse clusters, we apply K-means to the unified
representations and set the number of subclusters as
\begin{equation}
K_j=\max\left(2,\left\lfloor\frac{n_j}{100}\right\rfloor\right),
\label{eq:dynamic_k}
\end{equation}
thereby targeting approximately 100 instances per refined cluster. Specifically,
the refined subclusters are obtained by minimizing the objective 
\begin{equation}
\min_{\{\mathcal C_{jk},\boldsymbol\mu_{jk}\}_{k=1}^{K_j}}
\sum_{k=1}^{K_j}
\sum_{z_i\in\mathcal C_{jk}}
\left\|\mathbf h_i-\boldsymbol\mu_{jk}\right\|_2^2,
\label{eq:cluster_fission}
\end{equation}
where $\mathcal C_{jk}$ is the $k$-th refined subcluster of $\mathcal G_j$ and
$\boldsymbol\mu_{jk}$ is its centroid. This adaptive refinement produces
semantically more coherent candidate groups while avoiding a single
global clustering granularity for coarse clusters of different sizes.

Next, although memes may exhibit similar surface features or rhetorical patterns, their harmfulness can differ depending on the underlying semantic context. To prevent such samples with conflicting verdicts from being generalized into the same rule, we further partition each cluster according to the sample verdict $v_i$, yielding separate harmful and harmless candidate rules.
Finally, to balance rule generalization and long-tail risk coverage, we introduce an asymmetric support-based filtering strategy with $\tau_s=3$, where $s_j$ denotes the number of supporting instances for candidate rule:
\begin{equation}
\operatorname{Keep}(\mathcal{R}_j)=
\begin{cases}
1, & s_j \geq \tau_s,\\
1, & s_j < \tau_s \ \land\ v_j=\mathrm{Harmful},\\
0, & s_j < \tau_s \ \land\ v_j=\mathrm{Harmless}.
\end{cases}
\label{eq:asymmetric_pruning}
\end{equation}
Specifically, harmless rules are required to exhibit sufficient cross-instance recurrence to reduce overly instance-specific abstractions, whereas low-support harmful rules are retained to preserve potentially rare but important harmfulness patterns. This asymmetric design limits redundant rule proliferation while maintaining coverage of long-tail harmful scenarios.


\begin{figure*}[t] 
  \centering
  \includegraphics[width=\textwidth]{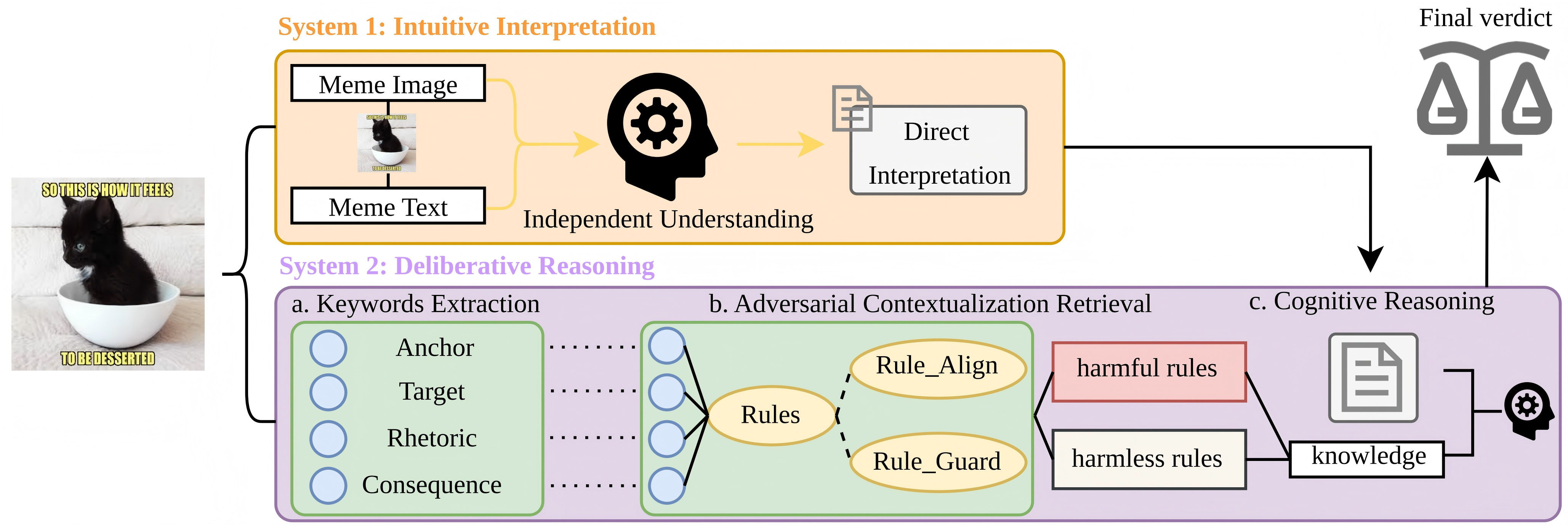} 
  \caption{Overview of Cognitive RAG. System 1 independently interprets the input meme without retrieved knowledge, while System 2 extracts preliminary semantic cues, retrieves adversarial harmful and harmless rules from the knowledge graph, and integrates the initial interpretation with retrieved evidence for knowledge-grounded reasoning and final harmfulness judgment.}
  \label{fig:rag}
\end{figure*}

\textbf{Harmfulness rule units.}
Although multimodal memes vary considerably in content and presentation, recurring harmfulness patterns often share a stable logical structure. We therefore canonicalize each retained cluster into a formalized rule unit:
\begin{equation}
\mathcal R_j
=
\left\langle
\mathcal M_j,\mathcal P_j,\mathcal S_j
\right\rangle,
\quad
\mathcal P_j=(A_j,Rh_j,T_j,C_j,V_j),
\label{eq:rule_unit}
\end{equation}
where $\mathcal M_j$ is a high-level semantic description of the harmfulness pattern, and $\mathcal P_j$ represents the stable attributes, including Anchor, Rhetoric, Target, Consequence and Verdict, which are obtained by abstracting logical primitives from the Corpus-Level Rule. $\mathcal S_j$ contains representative cases. Rather than relying on unconstrained knowledge generation, we employ a multimodal large language model to synthesize $\mathcal M_j$ and $\mathcal P_j$ based on the extracted logical primitives, supporting instances and harmfulness label within each cluster. This evidence-grounded generalization aims to capture recurring semantic patterns while reducing unsupported interpretations and preserving consistency with the underlying instances. Furthermore, each induced rule retains explicit associations with its supporting cases, facilitating knowledge traceability and subsequent retrieval-augmented reasoning.



\subsubsection{Topological Graph Modeling}

Harmfulness rule units extracted in the Formalized Rule Extraction stage are inherently high-order relational objects composed of multiple logical primitives. Conventional entity-relation graphs only encode binary facts as independent triples, which cannot natively match the structure of rule units. Such representation fragments the internal semantics of rules. Therefore, to fully retain the semantic integrity of each rule, realize explicit structured modeling, and facilitate retrieval and evidence tracing, we reconstruct plain-text harmfulness rule units into a structured harmfulness knowledge graph embedded with star-shaped topology, as shown in Harmfulness Knowledge Graph of Figure~\ref{fig:method}, which contains star-shaped rule subgraphs and cross-rule relations.

\textbf{Star-shaped rule subgraphs.}
For each harmfulness rule unit, we construct a rule-centered local star subgraph to explicitly organize its semantic attributes, supporting evidence, and associated verdict. The central Rule node represents a generalized harmfulness pattern and is connected to six types of peripheral nodes: Anchor (Anch.), Rhetoric (Rhet.), Target (Targ.), Consequence (Cons.), Support\_cases (Case), and Verdict (Verd.). The first four capture the key semantic characteristics underlying the rule, while Case nodes preserve supporting instances as empirical evidence and the Verd. node specifies the corresponding harmfulness judgment. The Mapping primitive is not explicitly included in the graph, as it mainly captures instance-specific cross-modal correspondences rather than reusable rule-level attributes. This rule-centered organization separates generalized harmfulness knowledge from its supporting instances while preserving their explicit associations.

We further define three types of structural relations to organize these dependencies. \textsc{Triggers} edges (Trig.) connect Anch., Rhet., Targ., and Cons. nodes to the Rule node, representing semantic conditions relevant to rule applicability. \textsc{Supports} edges (Supp.) connect Case nodes to their corresponding Rule node, linking generalized knowledge to its supporting instances. \textsc{Conclude} edges (Concl.) connect each Rule node to its Verd. node, associating the rule with its harmfulness judgment. Together, these relations establish explicit topological paths, providing a structured foundation for rule matching, evidence tracing, and knowledge retrieval.



\textbf{Cross-rule relations.}
Although rule-centered star subgraphs explicitly organize the semantic attributes and supporting evidence of individual rules, they lack direct associations across different rule units, limiting the exploration of related knowledge beyond local structures. To address this limitation, we introduce two types of cross-rule semantic relations based on semantic similarity and verdict polarity:
(1) \textsc{Semantically Aligned} edges (Align.) connect attribute nodes of the same semantic type when their similarity exceeds a threshold $\tau_a$. These relations capture semantic associations between attributes belonging to different rules, allowing retrieval to expand beyond directly matched attributes and identify relevant rules across different meme scenarios. This facilitates knowledge sharing and improves the accessibility of semantically related rule-level knowledge.
(2) \textsc{Guards Against} edges (Guard.) connect Rule nodes that exhibit high semantic similarity above a threshold $\tau_g$ but have opposite verdicts. These relations capture potentially contrasting harmfulness interpretations under similar semantic conditions, highlighting that semantically related patterns may lead to different judgments depending on their context. By linking rules with opposing verdicts, these edges facilitate the retrieval of contrastive evidence for examining rule applicability and distinguishing harmful from harmless interpretations.

Together, these cross-rule relations complement the rule-centered structural edges by enabling semantic retrieval expansion and polarity-aware knowledge exploration. The resulting graph integrates local rule-level evidence with cross-rule semantic associations, providing a structured knowledge foundation for subsequent Cognitive RAG.
Graph statistics are reported in Table~\ref{tab:graph_statistics}.


\subsection{Graph-Grounded Cognitive RAG}

\begin{table*}[t]
    \caption{Performance comparison on the MemeMind dataset. We compare MemeLogos with representative baselines. MemeLogos denotes the training-free framework, while MemeLogos\_Align further incorporates keyword and reasoning alignment. $^{\diamond}$ indicates direct evaluation of vision-language models without task-specific fine-tuning. \textbf{Bold} indicates the best result, while \underline{underlining} indicates the second-best result.}
      \label{experiment}
  \centering
  \small
  \renewcommand{\arraystretch}{1.5} 

  \begin{tabular*}{\textwidth}{@{\extracolsep{\fill}}llcccc}
    \toprule
    & \textbf{Method} & \textbf{Accuracy} & \textbf{Precision} & \textbf{Recall} & \textbf{F1-score} \\
    \midrule
    \multirow{2}{*}{Single-Modality} 
      & CLIP Text-Only~\cite{Radford2021LearningTV} & 74.14 & 75.54 & 56.92 & 64.92 \\
      & CLIP Image-Only~\cite{Radford2021LearningTV} & 80.25 & 79.47 & 71.74 & 75.41 \\
    \midrule
    \multirow{10}{*}{Multi-Modality} 
      & CLIP Image+Text~\cite{Radford2021LearningTV} & 80.81 & 80.81 & 71.06 & 75.62 \\
      & VisualBERT\_COCO~\cite{Visualbert} & 63.21 & 63.95 & 63.48 & 63.72 \\
      & Pro-Cap~\cite{Pro-Cap} & 75.58 & 78.09 & 78.76 & 78.44 \\
      & ISSUES~\cite{Issues} & 82.54 & 83.09 & 74.80 & 78.72 \\
      & MemeCLIP~\cite{MemeCLIP} & 79.18 & 80.48 & 78.98 & 79.72 \\
      & RGCL~\cite{Mei2023ImprovingHM} & 81.65 & 78.38 & 79.21 & 78.79 \\
      & RA-HMD~\cite{mei-etal-2025-robust} & 81.23 & 80.41 & 74.53 & 77.36 \\
      & Qwen2.5-VL-32B$^{\diamond}$~\cite{ref69} & 78.44 & 81.22 & 63.25 & 71.11 \\
      & Qwen3.6-plus$^{\diamond}$~\cite{qwen36plus} & 77.35 & \textbf{92.26} & 50.98 & 65.67 \\
      & GPT-5.5$^{\diamond}$~\cite{openai2026gpt55} & 77.91 & 86.97 & 57.37 & 69.14\\    
    \midrule
    \multirow{2}{*}{\textbf{Our Method}} 
   & \textbf{MemeLogos} & \underline{86.96} & 83.45 & \underline{87.07} & \underline{85.22} \\
    & \textbf{MemeLogos\_Align} & \textbf{89.50} & \underline{87.01} & \textbf{89.08} & \textbf{88.03} \\
    \bottomrule
  \end{tabular*}

\end{table*}

The harmfulness knowledge graph provides structured support for harmful meme detection. Building on this graph, we propose Cognitive RAG, as shown in Figure~\ref{fig:rag}, which supports Training-free inference to produce harmfulness analysis and enables Training-Alignment enhancement for meme feature extraction and reasoning.

\subsubsection{Training-free Cognitive RAG} 

Inspired by Dual-System Theory, where a fast system relies on instinct and experience while a  slow system drives conscious reasoning. we design Cognitive RAG with two complementary systems: intuitive interpretation (System 1) and deliberative reasoning (System 2). System 1 independently interprets the input meme without relying on retrieved knowledge, mitigating potential retrieval-induced bias. System 2 then performs deliberative reasoning through three stages: (1) core attribute extraction, which identifies preliminary semantic cues as retrieval hypotheses. (2) adversarial contextualization, which retrieves relevant rules to construct a polarity-balanced context containing competing interpretations. (3) cognitive reasoning, which integrates the initial interpretation with retrieved context to perform reasoning and hypothesis falsification, deriving the reasoning process and final verdict.

\textbf{System 1: Intuitive interpretation.}
Previous studies have shown that irrelevant or conflicting
retrieved knowledge can mislead model predictions and
interfere with the utilization of intrinsic knowledge~\cite{yu2024chain, amiraz2025distracting}. To mitigate this issue, System 1 independently interprets the input meme without accessing external knowledge, forming an initial rationale $h_i^{(1)}$ based solely on its multimodal content:
\begin{equation}
h_i^{(1)} \sim p_{\theta}(\cdot \mid x_i),
\label{eq:dual_system_reasoning_1}
\end{equation}
where $x_i$ denotes the input meme image and text.
This independent interpretation provides a complementary perspective for subsequent cognitive reasoning, helping the model integrate its intrinsic understanding of the meme with retrieved harmfulness knowledge.

\textbf{System 2: Deliberative reasoning.}
Building upon the System 1, System 2 incorporates retrieval knowledge to support harmfulness reasoning and verification. The deliberative process is organized into three stages: keyword extraction, adversarial contextualization retrieval, and cognitive reasoning.

First, in the keyword extraction stage, to reduce retrieval interference from redundant meme details, we extract a compact set of logical probes
$\mathcal{K}_i=\{\hat a_i,\hat r_i,\hat t_i,\hat c_i\}$,
corresponding to the anchor, rhetorical mechanism, target, and consequence of the input meme. These probes represent preliminary semantic hypotheses, providing structured cues for identifying potentially relevant rules. We then encode them into a retrieval query set $\mathbf{Q}$ for schema-aligned matching with the corresponding attribute nodes in the harmfulness knowledge graph.

Second, the adversarial contextualization retrieval stage provides more comprehensive semantic context by accounting for competing interpretations, as semantically similar patterns may correspond to different harmfulness verdicts. To this end, we construct a polarity-balanced context through graph-grounded retrieval. Specifically, we first compute the semantic relevance between $\mathbf{Q}$ and rule attribute nodes. We obtain directly matched rules $\mathcal{R}_{\mathrm{dir}}$ through \textsc{Triggers} edges, expand semantically related rules $\mathcal{R}_{\mathrm{rel}}$ through \textsc{Aligned} edges, and identify contrastive rules $\mathcal{R}_{\mathrm{risk}}$ through \textsc{Guards} edges. Candidate rules are then lexicographically reranked according to the priority
$\mathcal{R}_{\mathrm{dir}} >
\mathcal{R}_{\mathrm{rel}} >
\mathcal{R}_{\mathrm{risk}}$. Finally, we retrieve Top-$K$ harmful and Top-$K$ harmless rules to construct the adversarial context $\Sigma$, providing competing rule-level knowledge for subsequent reasoning.

Finally, in the cognitive reasoning stage, given the initial interpretation $h_i^{(1)}$ and the retrieved adversarial context $\Sigma$, System 2 performs knowledge-grounded reasoning to critically examine against harmful and harmless rules. By jointly considering the initial interpretation and retrieved knowledge, the model conducts comprehensive reasoning to produce a refined rationale $h_i^{(2)}$ and the final verdict $\hat y_i$:
\begin{equation}
(\hat y_i,h_i^{(2)})
\sim p_{\theta}\left(
\cdot \mid x_i,h_i^{(1)},\Sigma
\right),
\label{eq:dual_system_reasoning_2}
\end{equation}
This dual-system design combines independent multimodal interpretation with contrastive knowledge-grounded reasoning, mitigating potential retrieval-induced bias while supporting more reliable and traceable harmfulness judgments.

\subsubsection{Training-Alignment Cognitive RAG}

To further improve the quality of knowledge retrieval and reasoning, we develop MemeLogos\_Align, a training-aligned variant of Cognitive RAG through two optimization strategies: keywords alignment and reasoning alignment.

\textbf{Keywords alignment.}
To enhance the consistency of core semantic attribute extraction, we perform supervised fine-tuning using logical primitives derived from annotated rules as supervision targets. This process guides the model to map multimodal meme inputs into structured semantic representations aligned with the attribute schema of the harmfulness knowledge graph, rather than unconstrained textual descriptions. By reducing the semantic mismatch between extracted queries and graph attributes, keyword alignment aims to improve retrieval precision and the relevance of retrieved rules.

\textbf{Reasoning alignment.}
To further improve the utilization of retrieved knowledge during cognitive reasoning, we employ Group Relative Policy Optimization (GRPO)~\cite{GRPO} to optimize the model through outcome-based feedback. For each meme, the model samples a group of candidate reasoning trajectories, each producing a final harmfulness verdict. We assign an accuracy-based reward according to the correctness of the predicted verdict:
\begin{equation}
r_{i,g}
=
\mathbb I[\hat y_{i,g}=y_i],
\label{eq:accuracy_reward}
\end{equation}
where $\hat y_{i,g}$ denotes the verdict produced by the $g$-th trajectory and $y_i$ is the ground-truth label. Through group-relative optimization, this reward encourages reasoning trajectories that lead to accurate harmfulness judgments, facilitating more effective integration of independent semantic interpretations and retrieved rule-level knowledge.

\section{Experiment}
\subsection{Experimental Setup}

\subsubsection{Baselines and Datasets} The experiments are divided into two groups. The single-modality group adopts the CLIP~\cite{Radford2021LearningTV} to process text and image modalities independently, while the multimodality group includes representative multimodal methods, such as CLIP, Pro-Cap~\cite{Pro-Cap}, ISSUES~\cite{Issues}, MemeCLIP~\cite{MemeCLIP}, VisualBERT\_COCO~\cite{Visualbert}, U-CoT+~\cite{Pan2025ReadAY}, RGCL~\cite{Mei2023ImprovingHM}, RA-HMD~\cite{mei-etal-2025-robust}, ExPO-HM~\cite{Mei2025ExPOHMLT}, and vision-language models. Except for the vision-language models baselines, all methods in Table~\ref{experiment} are uniformly trained and evaluated on MemeMind. We also conduct cross domain experiments in PrideMM~\cite{MemeCLIP}.

\subsubsection{Implementation Details} We adopt Qwen3-VL-8B~\cite{Bai2025Qwen3VLTR} as the backbone for Cognitive RAG inference and Qwen3-VL-32B~\cite{Bai2025Qwen3VLTR} for knowledge extraction and rule generalization. All-Mpnet-Base-V2 as the semantic encoder for knowledge construction and retrieval. For formalized rule extraction, the extracted logical primitives are encoded into 768-dimensional embeddings with a batch size of 128. We first perform coarse clustering based on rhetorical mechanisms using a cosine similarity threshold of $\tau_r=0.7$. Within each coarse cluster, the Anchor, Mapping, Target, and Consequence primitives are concatenated and encoded for K-Means refinement, with the number of clusters adaptively determined to contain approximately 100 samples per refined cluster. Harmful and harmless samples are subsequently separated according to their verdicts to avoid merging contradictory patterns. For refined clusters containing more than 10 samples, K-Means-based diversity sampling is employed to select representative supporting cases. For topological graph modeling, we set the similarity thresholds of \textsc{Aligned} and \textsc{Guards} edges to $\tau_a=0.9$ and $\tau_g=0.75$, respectively, and construct the resulting harmfulness knowledge graph using NetworkX. During Cognitive RAG inference, we retrieve the top-3 rules from each polarity for contrastive reasoning. To prevent data leakage, all rule extraction and knowledge graph construction are performed exclusively on the training split of MemeMind. We report Accuracy, Precision, Recall, and F1-score as the primary evaluation metrics.

\begin{table}[t]
\caption{Cross-domain zero-shot evaluation of MemeLogos on PrideMM.}
\label{tab:cross-domain}
\centering
\small
\setlength{\tabcolsep}{3pt}

\begin{tabular*}{\columnwidth}{@{\extracolsep{\fill}}lcc@{}}
\toprule
\textbf{Method} & \textbf{Accuracy} & \textbf{F1} \\
\midrule

\multicolumn{3}{c}{\textbf{In-domain Supervised Evaluation}}\\
\midrule
CLIP      & 72.39 & 72.33 \\
MemeCLIP  & 76.06 & 75.09 \\
U-CoT+    & 71.60 & 71.37 \\
RGCL      & 76.30 & 76.50 \\
RA-HMD    & \textbf{78.10} & 78.40 \\
ExPO-HM   & --- & 78.70 \\

\midrule
\multicolumn{3}{c}{\textbf{Cross-domain Zero-shot Evaluation}}\\
\midrule
Qwen3-VL-8B & 64.69 & 69.61 \\
MemeLogos   & 77.51 & \textbf{79.50} \\

\bottomrule
\end{tabular*}
\end{table}

\subsection{Experimental Results}
Table~\ref{experiment} reports the main results on MemeMind. 
Among single-modality baselines, the image-only CLIP model substantially outperforms its text-only counterpart, achieving an F1-score of 75.41 compared with 64.92, which indicates that visual information provides important cues for harmful meme detection. However, combining image and text with CLIP yields only a marginal improvement to 75.62 F1, suggesting that simple multimodal fusion is insufficient to capture the implicit semantic interactions underlying harmfulness. More advanced multimodal approaches, including Pro-Cap, ISSUES, MemeCLIP, RGCL, and RA-HMD, achieve stronger overall performance, confirming the effectiveness of cross-modal modeling and task-specific semantic enhancement. Nevertheless, their performance remains limited when harmfulness depends on implicit contextual associations and reasoning beyond surface-level visual-textual correspondence. In particular, the strongest existing baseline achieves 82.54\% accuracy, leaving considerable room for further improvement.

In addition, stronger general vision-language models do not necessarily achieve superior performance on harmful meme detection under direct zero-shot evaluation. As shown in Table~\ref{experiment}, Qwen2.5-VL-32B, Qwen3.6-plus, and GPT-5.5 obtain F1-scores of 71.11, 65.67, and 69.14, respectively. In particular, Qwen3.6-plus and GPT-5.5 achieve relatively high precision but substantially lower recall, indicating a conservative prediction tendency that misses a considerable proportion of harmful memes. These results suggest that general multimodal capability alone is insufficient to reliably capture the implicit contextual associations and compositional harmfulness semantics required by this task. Task-specific harmfulness knowledge and structured reasoning therefore remain important for improving both recognition coverage and decision reliability.

In contrast, MemeLogos achieves 86.96\% accuracy and 85.22\% F1, outperforming the strongest existing baselines. These improvements demonstrate the effectiveness of jointly leveraging structured harmfulness knowledge and Cognitive RAG. The harmfulness knowledge graph provides reusable and organized task-specific knowledge, while Cognitive RAG enables the model to effectively incorporate such knowledge into the reasoning process, thereby improving its ability to capture implicit harmful semantics and make more reliable predictions.
Furthermore, MemeLogos\_Align improves the performance to 89.50\% accuracy and 88.03\% F1, yielding additional gains of 2.54 and 2.81 percentage points over the training-free MemeLogos, respectively. This further demonstrates that aligning both semantic keyword extraction and reasoning behavior with the structured harmfulness knowledge can enhance the utilization of retrieved evidence and improve the overall detection capability.

Having demonstrated the strong in-domain performance of MemeLogos on MemeMind, we further evaluate the cross-domain generalization ability of MemeLogos. Specially, we conduct zero-shot transfer experiments on PrideMM using only the harmfulness knowledge constructed from the MemeMind training set. As shown in Table~\ref{tab:cross-domain}, we include the reported results of existing in-domain supervised methods as references. MemeLogos still maintains stable detection performance on PrideMM, indicating that the learned rule-level harmfulness knowledge captures reusable patterns that generalize across different data distributions.

\begin{table}[t]
  \caption{Computational efficiency of different knowledge representations. Access denotes knowledge retrieval, while inference denotes
    downstream prediction without full reasoning.}
  \label{tab:retrieval_efficiency}
  \centering
  \small
  \renewcommand{\arraystretch}{1.15}
  \setlength{\tabcolsep}{4pt}

  \begin{tabular*}{\columnwidth}{
    @{\extracolsep{\fill}}
    lccc
    @{}
  }
    \toprule
    \textbf{Method}
      & \textbf{Access (s)}
      & \textbf{Inference (s)}
      & \textbf{Total (s)} \\
    \midrule

    Rule-RAG
      &0.076
      &0.092
      &0.168 \\

    Graph-RAG
      & 0.170
      & 0.107
      & 0.277 \\

    Instance-RAG
      & 0.139
      & 0.278
      & 0.417 \\

    \bottomrule
  \end{tabular*}
\vspace{-10pt}
\end{table}

\begin{table}[t]
  \caption{Ablation on graph structure.}
  \label{tab:graph_ablation}
  \centering
  \small
  \renewcommand{\arraystretch}{1.2}
  \setlength{\tabcolsep}{4pt}

  \begin{tabular*}{\columnwidth}{
    @{\extracolsep{\fill}}
    >{\centering\arraybackslash}p{1.8cm}
    lcc
    @{}
  }
    \toprule
    \textbf{Category}
      & \textbf{Model Variant}
      & \textbf{Accuracy}
      & \textbf{F1} \\
    \midrule

    \multirow{2}{*}{\textbf{Representation}}
      & Instance-RAG
      & 82.96
      & 80.44 \\
      & Rule-RAG
      & 82.76
      & 80.41 \\

    \midrule

    \multirow{2}{*}{\textbf{Variants}}
      & w/o \textsc{Aligned} edges
      & 86.13
      & 84.47 \\
      & w/o \textsc{Guards} edges
      & 85.32
      & 83.51 \\

    \midrule

    \textbf{Our Method}
      & MemeLogos
      & \textbf{86.96}
      & \textbf{85.22} \\

    \bottomrule
  \end{tabular*}

\end{table}

\begin{table}[t]
  \caption{Ablation of components in Cognitive RAG.}
  \label{tab:rag_ablation_full_text}
  \centering
  \small
  \setlength{\tabcolsep}{0pt} 
  \renewcommand{\arraystretch}{1.2}

  \begin{tabular*}{\columnwidth}{@{\extracolsep{\fill}}lcc}
    \toprule
    \textbf{Variant} & \textbf{Accuracy} & \textbf{F1} \\
    \midrule
    w/o Keywords Extraction      & 83.49 & 83.13 \\
    w/o Adversarial Context      & 85.11 & 82.64 \\
    w/o Cognitive Reasoning      & 64.55 & 39.47 \\
    \midrule
    \textbf{MemeLogos} & \textbf{86.96} & \textbf{85.22} \\
    \bottomrule
  \end{tabular*}

\end{table}

\subsection{Ablation Study}
In this section, we evaluate the contribution of each component design in MemeLogos. For all ablation experiments, we maintain the same experimental configurations described in previous sections.

\subsubsection{Ablation on Knowledge Graph}
To independently analyze the effects of knowledge representation and graph structure, we conduct ablation experiments with all other experimental configurations fixed. We first compare the controlled computational overhead of different knowledge access strategies in Table~\ref{tab:retrieval_efficiency}, while the actual inference overhead of the complete MemeLogos framework is detailed in Appendix B. Specifically, Instance-RAG directly retrieves similar meme instances, Rule-RAG retrieves induced rules based on plain-text semantic similarity, and Graph-RAG adopts the knowledge graph-based rule retrieval strategy used in MemeLogos. To isolate the computational cost of knowledge access, all variants use the same direct prediction setting rather than the full end-to-end MemeLogos pipeline. Under this controlled setting, the total latency is 0.417s, 0.168s, and 0.277s for Instance-RAG, Rule-RAG, and Graph-RAG, respectively.

Then, we further evaluate the variants and MemeLogos under the full end-to-end inference setting as shown in Table~\ref{tab:graph_ablation}. The results show that retrieving induced rules in the plain-text space achieves performance comparable to direct retrieval of similar meme instances. This indicates that the induced rule-level knowledge can effectively preserve harmfulness-relevant information from meme instances. However, Rule-RAG still underperforms MemeLogos with the star-shaped graph structure, demonstrating the benefit of structured knowledge organization for retrieval and reasoning. 

Moreover, removing either \textsc{Aligned} or \textsc{Guards} edges leads to a consistent yet relatively limited performance degradation. This indicates that the semantic cross-rule relations do provide useful signals for expanding relevant knowledge and characterizing semantic associations and decision boundaries among related rules. However, their contribution is smaller than that of the structural relations directly induced by the rule schema. This suggests that the primary benefit of the graph comes from its explicit rule-centered structural organization, while semantic relation edges mainly serve as complementary cues for relational expansion and contrastive retrieval.

\begin{table}[t]
  \caption{Performance of MemeLogos across different multimodal backbones.}
  \label{tab:extensibility}
  \centering
  \small
  \renewcommand{\arraystretch}{1.10}

  \begin{tabular*}{\columnwidth}{
    @{\extracolsep{\fill}}
    l
    l
    c
    c
    @{}
  }
    \toprule
    \textbf{Backbone}
      & \textbf{Variant}
      & \textbf{Accuracy}
      & \textbf{F1} \\
    \midrule

    \multirow{2}{*}{Qwen3-VL-8B}
      & \textit{Zero-shot}
      & 73.24
      & 66.32 \\
      & \textit{MemeLogos}
      & \textbf{86.96}
      & \textbf{85.22} \\

    \midrule

    \multirow{2}{*}{InternVL3.5-8B}
      & \textit{Zero-shot}
      & 69.75
      & 63.81 \\
      & \textit{MemeLogos}
      & \textbf{85.43}
      & \textbf{84.76} \\

    \midrule

    \multirow{2}{*}{MiniCPM-V-4.5}
      & \textit{Zero-shot}
      & 75.62
      & 75.67 \\
      & \textit{MemeLogos}
      & \textbf{81.95}
      & \textbf{82.54} \\
    \bottomrule
  \end{tabular*}

\end{table}

\subsubsection{Ablation on Component in Cognitive RAG}
To independently evaluate the contribution of core components in cognitive RAG, we conduct ablation experiments on MemeLogos with the training-free settings and other configurations consistent. As shown in Table~\ref{tab:rag_ablation_full_text}, we first replace the keywords extraction module with a free-form caption or query automatically generated by the model for retrieval. The results show that replacing the structured logical keywords with such free-form queries leads to a performance degradation. This suggests that explicitly aligning retrieval queries with the graph schema provides more discriminative and task-relevant cues than unconstrained textual descriptions, thereby improving rule retrieval. 

We then replace adversarial contextualization with conventional Top-\(K\) rule retrieval, where rules are solely according to retrieval relevance. The resulting performance degradation suggests that polarity-aware contrastive context helps the model distinguish semantically similar but decisionally different cases, leading to more reliable harmfulness judgments.

In addition, with other settings fixed, we replace Cognitive Reasoning with a standard retrieval-augmented inference scheme. The results show that
removing Cognitive Reasoning substantially degrades performance, showing that simply exposing the model to the conflict retrieved knowledge may introduce redundant or distracting information and consequently impair prediction. In contrast, the dual-system cognitive reasoning process explicitly verifies, contrasts, and refines the retrieved evidence before making the final decision, enabling more reliable knowledge utilization.

\subsubsection{Ablation on Backbones}
To evaluate the extensibility of MemeLogos, we conduct comprehensive experiments across multiple backbones, with other settings consistent. To avoid architectural overlap, we select MiniCPM-V-4.5~\cite{Yu2025MiniCPMV4C} and InternVL3.5-8B~\cite{Wang2025InternVL35AO} as backbones and use the same zero-shot CoT prompt. As illustrated in Table~\ref{tab:extensibility}, MemeLogos consistently improves the performance of other backbones, validating the robustness of MemeLogos as a model-agnostic framework for harmful meme reasoning and detection.

\subsubsection{Ablation on Training Alignment} 
To isolate and evaluate the impact of Keyword Alignment and Reasoning Alignment, we conduct ablation experiments by keeping other components fixed. As shown in Table~\ref{tab:training_ablation_simple}, MemeLogos denotes the training‑free setting, while other strategy represents its trained variant, the inclusion of Keyword Alignment yields substantial improvements across multiple core metrics, demonstrating that Keyword Alignment is crucial for enhancing the precision of semantic retrieval within the multimodal context. Furthermore, the performance uplift from Reasoning Alignment validates the efficacy of reasoning refinement via GRPO. By facilitating reasoning exploration, this process enables the model to better utilize the retrieved structured knowledge, thereby improving prediction reliability in complex cases.

\subsubsection{Parameter Sensitivity Analysis} To investigate the effects of key hyperparameters on MemeLogos, we perform parameter sensitivity analyses for the harmfulness knowledge graph and Cognitive RAG in MemeLogos on the MemeMind dataset. The remaining training hyperparameters were fixed without tuning based on the default configurations of the corresponding backbone and optimization framework.

\textbf{Graph sensitivity analysis.}
We investigate the sensitivity of MemeLogos to the key
parameters involved in rule induction and graph construction. Specifically,
we analyze the coarse clustering threshold $\tau_r$, the Semantically Aligned edge 
threshold $\tau_a$, and the Guards Against edge threshold $\tau_g$. All configurations are evaluated using the same fixed set of 5,000 training instances and 200 validation instances.

\begin{figure}[t]
    \centering
    \includegraphics[width=\columnwidth]{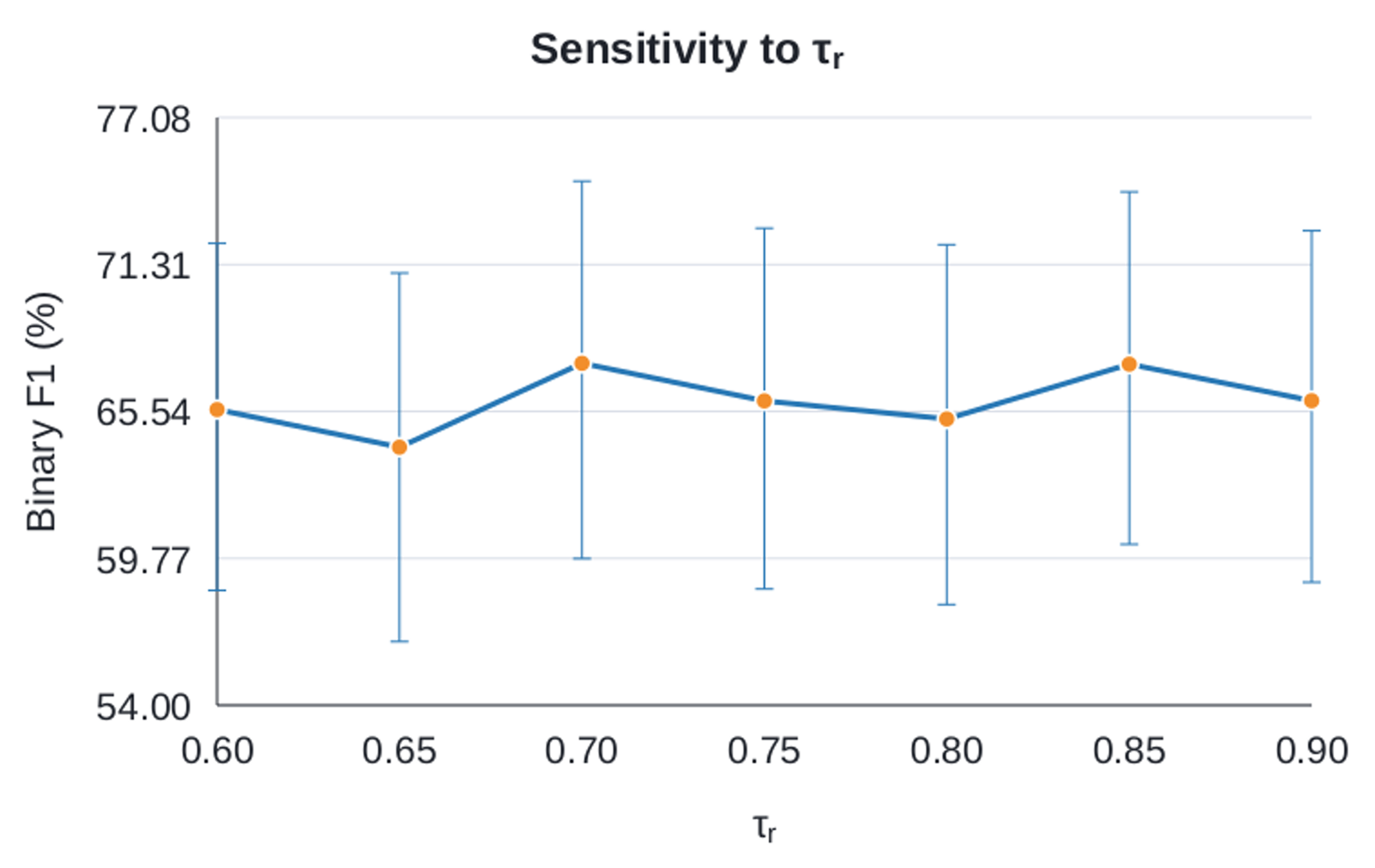}
    \caption{The sensitivity experiments of MemeLogos to the rule-clustering threshold
    $\tau_r$ conducted on a fixed 5,000/200 diagnostic subset. Error bars denote 95\% confidence intervals estimated using
5,000 class-stratified bootstrap resamples.}
    \label{fig:tau_r_sensitivity}
\end{figure}

For $\tau_r$, we vary the threshold from $0.60$ to $0.90$ while keeping
the graph parameters fixed. As shown in Fig~\ref{fig:tau_r_sensitivity},
the performance exhibits a
non-monotonic but relatively stable trend across different values of
$\tau_r$. The best observed overall performance is obtained at
$\tau_r=0.70$, achieving an Accuracy of 71.50\% and an F1-score of
67.40\%. Comparable performance is also observed at several other
settings, indicating that the method is not highly sensitive to moderate changes in the rule-clustering threshold.

\begin{figure}[t]
    \centering
    \includegraphics[width=\columnwidth]{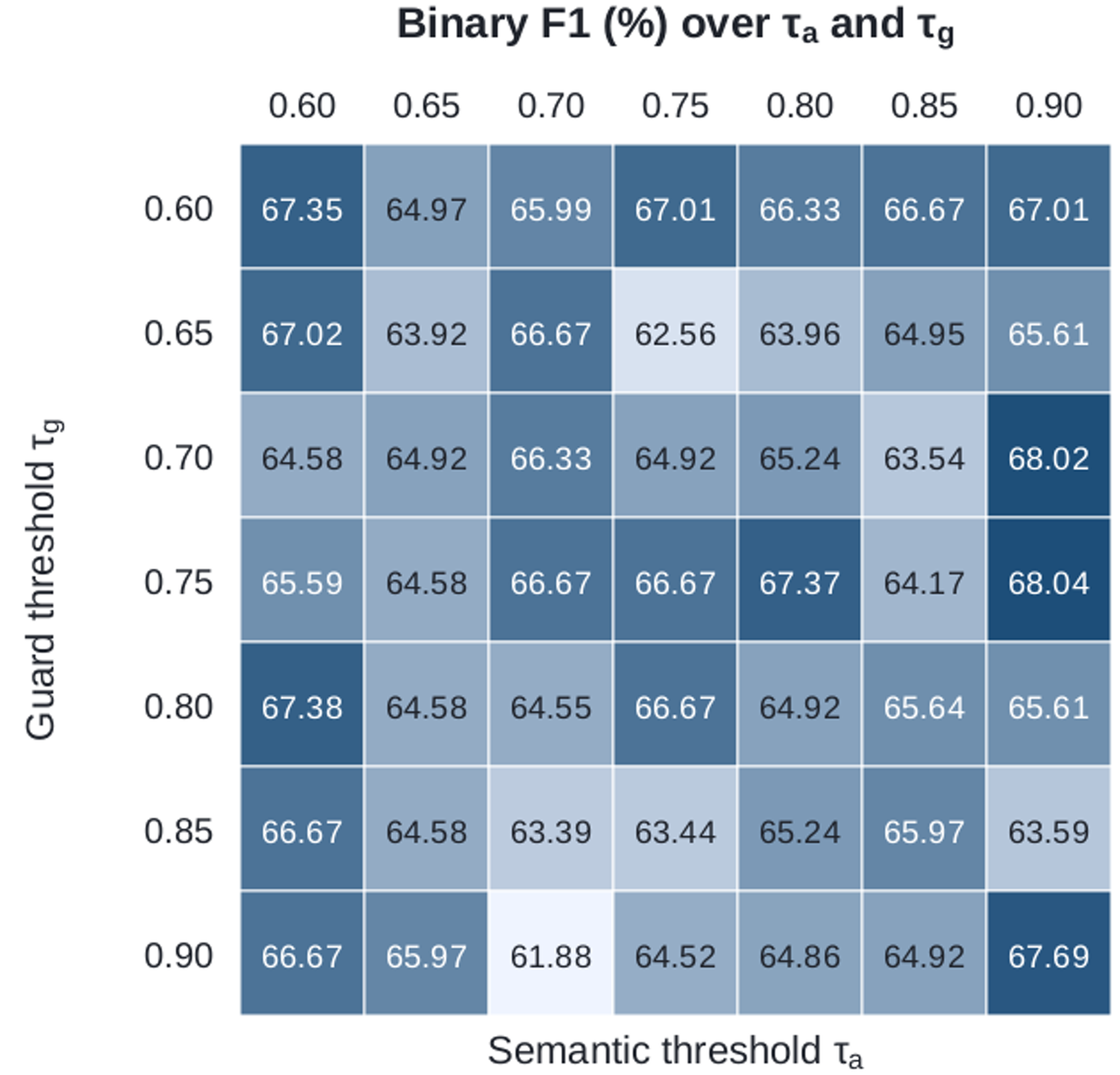}
    \caption{The sensitivity experiments of MemeLogos to the combinations of the Semantically Aligned edge
    threshold $\tau_a$ and Guards Against edge  threshold $\tau_g$ conducted on a fixed 5,000/200 diagnostic subset.}
    \label{fig:tau_ag_sensitivity}
\end{figure}

We further jointly vary $\tau_a$ and $\tau_g$ from 0.60 to 0.90, as shown
in Fig~\ref{fig:tau_ag_sensitivity}. The highest observed
F1-score of 68.04\% is achieved at
$(\tau_a,\tau_g)=(0.90,0.75)$, while several other parameter
combinations obtain similar performance. The absence of a single sharp
performance peak suggests that MemeLogos does not rely on a narrowly
tuned graph-construction setting and maintains relatively stable
performance over a broad range of threshold combinations.

\textbf{Retrieval count sensitivity analysis.}
We further analyze the sensitivity of Cognitive RAG to the number of
retrieved rules. Specifically, we vary the number of rules retrieved
for each polarity, denoted by $K$, while keeping all other settings
unchanged. The results are reported in Table~\ref{tab:retrieval_count}.

As shown in Table~\ref{tab:retrieval_count}, the model achieves comparable performance across different values of $K$, with an accuracy variation of only 0.34 percentage points. Retrieving too few rules may limit the availability of relevant evidence, whereas retrieving more rules may introduce additional irrelevant or redundant information. The best result is obtained at $K=3$, suggesting that a moderate retrieval size provides a suitable balance between evidence coverage and retrieval noise. Nevertheless, the limited performance variation demonstrates that Cognitive RAG is robust to the number of rules retrieved for each polarity.

\begin{table}[t]
  \caption{Sensitivity analysis of the number of retrieved rules for
  each polarity.}
  \label{tab:retrieval_count}
  \centering
  \small
  \renewcommand{\arraystretch}{1.15}

  \begin{tabular*}{\columnwidth}{@{\extracolsep{\fill}}ccc@{}}
    \toprule
    \textbf{Rules per Polarity ($k$)}
      & \textbf{Accuracy} & \textbf{F1}\\
    \midrule
    1 & 86.65 & 84.93\\
    3 & \textbf{86.96} & \textbf{85.22}\\
    5 & 86.62 & 84.91\\
    \bottomrule
  \end{tabular*}
\end{table}

\begin{table}[t]
    \caption{Ablation of Training Alignment.}
      \label{tab:training_ablation_simple}
  \centering
  \small
  \setlength{\tabcolsep}{0pt} 
  \renewcommand{\arraystretch}{1.2}
  
  \begin{tabular*}{\columnwidth}{@{\extracolsep{\fill}}lcc}
    \toprule
    \textbf{Strategy} & \textbf{Accuracy} & \textbf{F1} \\
    \midrule
    MemeLogos            & 86.96 & 85.22 \\
    \midrule
    Keywords Alignment    & 87.20 & 85.68 \\
    Reasoning Alignment      & 88.57 & 86.88 \\
    \midrule
    \textbf{MemeLogos\_Align} & \textbf{89.50} & \textbf{88.03} \\
    \bottomrule
  \end{tabular*}

\end{table}

\begin{figure*}[t]
  \centering
  \small
  
  \tcbset{
    casebox/.style={colback=white, colframe=gray!30, fonttitle=\bfseries, arc=2pt, outer arc=2pt, boxsep=3pt, left=4pt, right=4pt, top=4pt, bottom=4pt},
    rulebox/.style={colback=gray!5, colframe=gray!20, arc=2pt, left=4pt, right=4pt, top=4pt, bottom=4pt, boxrule=0.5pt},
    intuitionbox/.style={colback=blue!5, colframe=blue!20, arc=2pt, left=4pt, right=4pt, top=4pt, bottom=4pt, boxrule=0.5pt},
    auditingbox/.style={colback=orange!5, colframe=orange!20, arc=2pt, left=4pt, right=4pt, top=4pt, bottom=4pt, boxrule=0.5pt}
  }

  \begin{tcolorbox}[colframe=black!10, colback=white, arc=3pt, boxsep=2pt, left=2pt, right=2pt, top=2pt, bottom=2pt]
    \begin{tabularx}{\linewidth}{@{} p{0.35\textwidth} | X @{}}
      
      \begin{minipage}[t]{0.35\textwidth}
        \vspace{0pt}
        \centering
        \includegraphics[width=0.8\linewidth]{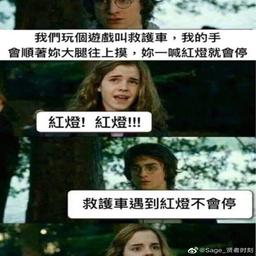}
        \par\vspace{4pt}
        \raggedright
        {\sffamily\bfseries Expert Reference}
        \begin{itemize}[leftmargin=12pt, nosep, itemsep=3pt, topsep=4pt]
          \item Sexual consent trivialized via "game" rule.
        \end{itemize}
        
        \raggedright
        {\sffamily\bfseries{Logical Keywords}}
        \begin{itemize}[leftmargin=12pt, nosep, itemsep=3pt, topsep=1pt]
          \item \textbf{Anchor}: Ambulance emergency privilege.
          \item \textbf{Rhetoric}: False Equivalence.
          \item \textbf{Mapping}: sexualized game.
          \item \textbf{Consequence}: Non-consensual sexualization.
          \item \textbf{Target}: Personal Consent Norms.
        \end{itemize}
        
      \end{minipage}

      & 

      \begin{minipage}[t]{\linewidth}
        \vspace{10pt}

        {\sffamily\bfseries{Rule Reference}}
        \begin{tcolorbox}[rulebox]
          \texttt{RULE\_CONSENT\_01\_HARMFUL}: 
          \textit{Normalizing non-consensual or sexual behavior through humor or gamified contexts, thereby creating harmful patterns.}
        \end{tcolorbox}
        \vspace{10pt}
        {\sffamily\bfseries System 1: Endogenous Intuition}
        \begin{tcolorbox}[intuitionbox]
         This meme extracts romantic couple scenes in films and embeds text related to traffic rules, constructing a playful rule‑breaking scenario via a humorous context.
        \end{tcolorbox}
        \vspace{10pt}
        {\sffamily\bfseries System 2: Rule-Guided Verification}
        \begin{tcolorbox}[auditingbox]
          The core mechanism — using a game rule to mock sexual consent — aligns with the harmful pattern in \texttt{RULE\_CONSENT\_01}. It trivializes consent boundaries and undermines gender norms, functionally equivalent to decontextualized mocking. The remaining rules mostly correspond to banter and helplessness in daily life. They do not match this case at the pattern level and exhibit target mismatch, hence being inapplicable. Consequently, this meme is harmful.
        \end{tcolorbox}
        \vspace{10pt}
        \begin{tcolorbox}[colback=black!80, colframe=black, boxsep=2pt, left=6pt, right=6pt, top=4pt, bottom=4pt]
          \textcolor{white}{\textbf{Final Verdict}:} \hfill \textcolor{red!40}{\textbf{\textsc{Harmful}}} \hfill \textcolor{white}{\textbf{Confidence}: \texttt{0.982}}
        \end{tcolorbox}
      \end{minipage}

    \end{tabularx}
  \end{tcolorbox}

  \caption{Qualitative illustration of the reasoning process in MemeLogos. The framework extracts key semantic cues, retrieves relevant harmfulness rules, and performs dual-system reasoning to progressively verify the evidence and derive the final harmfulness verdict.}
  \label{fig:qualitative_audit}
\end{figure*}

\subsection{Interpretability Analysis}

We conduct a human evaluation on 1,000 randomly sampled memes using a 5-point Likert scale. Three independent annotators assess the generated reasoning along three complementary dimensions: \textbf{Faithfulness} (Faith.), which measures the consistency between the generated reasoning and the observable multimodal evidence in the input meme. A high score requires that the explanation correctly reflects the visual and textual content without introducing unsupported entities, intentions, or semantic relations. We emphasize that this metric evaluates input-grounded consistency rather than causal faithfulness between the explanation and the model's internal decision process; \textbf{Sufficiency} (Suff.), which evaluates whether the explanation covers the key evidence and reasoning steps required to understand the final prediction. A sufficient explanation should identify the critical multimodal cues, clarify their semantic interaction, and explicitly connect them to the harmfulness judgment. Explanations that merely restate the meme content or provide a verdict without intermediate justification receive lower scores; and \textbf{Rule Relevance} (Rel.), which evaluates whether the retrieved rule-level knowledge is semantically applicable to the current meme and whether the generated reasoning uses such knowledge appropriately. A high score requires that the retrieved rules capture the central harmfulness mechanism of the meme rather than only matching superficial visual or lexical features. This metric is only applicable to MemeLogos, since the vanilla backbone does not perform rule retrieval. These metrics respectively evaluate whether the reasoning remains grounded in the input meme, whether it provides sufficient evidence for understanding the final judgment, and whether the retrieved harmfulness rules are applicable to the current case. Detailed information of the three dimensions is shown in Appendix C.

Table~\ref{tab:human_eval} shows that MemeLogos achieves higher Faithfulness and Sufficiency scores than the Qwen3-VL-8B backbone with CoT prompting, while also obtaining a high Rule Relevance score. The three annotators exhibit substantial agreement, with ordinal Krippendorff's $\alpha$ values of 0.79, 0.83, and 0.74 for Faithfulness, Sufficiency, and Rule Relevance, respectively. These results indicate that the retrieved rule-level knowledge provides effective support for generating more grounded and sufficiently justified reasoning, thereby enhancing the interpretability of harmfulness judgments. 

To further illustrate how this knowledge-grounded reasoning process operates in practice, we provide a representative qualitative example in Figure~\ref{fig:qualitative_audit}. In this example, the meme combines the visual context of an ambulance-related emergency privilege with a sexualized “game-rule” interpretation, where the humorous framing implicitly trivializes personal consent. MemeLogos first identifies the key multimodal cues and abstracts them into structured logical keywords, including the anchor, false-equivalence rhetoric, sexualized mapping, target, and potential consequence. Based on these cues, the framework retrieves a relevant harmfulness rule concerning the normalization of non-consensual or sexual behavior through humorous or gamified contexts. System~1 then forms an initial interpretation of the meme, while System~2 contrasts the retrieved rule with alternative candidates and verifies its applicability before producing the final harmful verdict.

\begin{figure*}[t]
  \centering
  \small
  
  \tcbset{
    casebox/.style={colback=white, colframe=gray!30, fonttitle=\bfseries, arc=2pt, outer arc=2pt, boxsep=3pt, left=4pt, right=4pt, top=4pt, bottom=4pt},
    rulebox/.style={colback=gray!5, colframe=gray!20, arc=2pt, left=4pt, right=4pt, top=4pt, bottom=4pt, boxrule=0.5pt},
    intuitionbox/.style={colback=blue!5, colframe=blue!20, arc=2pt, left=4pt, right=4pt, top=4pt, bottom=4pt, boxrule=0.5pt},
    auditingbox/.style={colback=orange!5, colframe=orange!20, arc=2pt, left=4pt, right=4pt, top=4pt, bottom=4pt, boxrule=0.5pt}
  }

  \begin{tcolorbox}[colframe=black!10, colback=white, arc=3pt, boxsep=2pt, left=2pt, right=2pt, top=2pt, bottom=2pt]
    \begin{tabularx}{\linewidth}{@{} p{0.35\textwidth} | X @{}}
      
      \begin{minipage}[t]{0.35\textwidth}
        \vspace{0pt}
        \centering
        \includegraphics[width=0.95\linewidth]{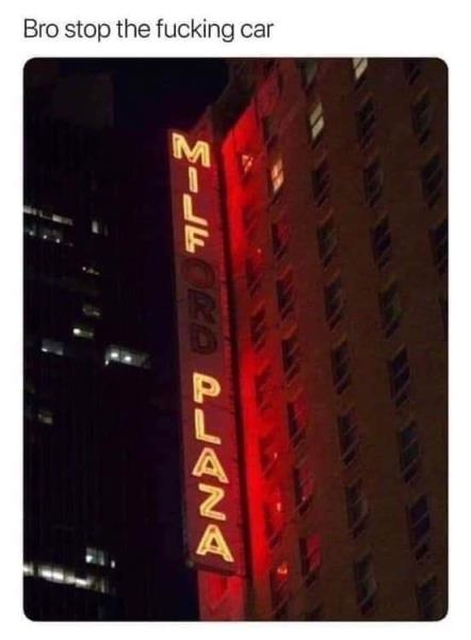}
        \par\vspace{4pt}
        \raggedright
        {\sffamily\bfseries Expert Reference}
        \begin{itemize}[leftmargin=12pt, nosep, itemsep=3pt, topsep=4pt]
          \item Sexualize and objectify women through vulgar slang.
        \end{itemize}
        
        \raggedright
        {\sffamily\bfseries{Logical Primitives}}
        \begin{itemize}[leftmargin=12pt, nosep, itemsep=3pt, topsep=1pt]
          \item \textbf{\scriptsize Anchor}: Illuminated hotel sign.
          \item \textbf{\scriptsize Rhetoric}: Visual juxtaposition.
          \item \textbf{\scriptsize mapping}: sexualized nomenclature.
          \item \textbf{\scriptsize consequence}: Public order and good morals.
          \item \textbf{\scriptsize Target}: Personal Public building.
        \end{itemize}
        
      \end{minipage}

      & 

      \begin{minipage}[t]{\linewidth}
        \vspace{10pt}

        {\sffamily\bfseries{Rule Reference}}
        \begin{tcolorbox}[rulebox]
          \texttt{ RULE\_JUXTAPOSITION\_EXPRESSIONS\_01\_HARMLESS}: 
          \textit{Harmless use of visual juxtaposition to highlight emotional or situational contrasts through relatable, humorous, or satirical expressions without malicious intent.}
        \end{tcolorbox}
        \vspace{10pt}
        {\sffamily\bfseries System 1: Endogenous Intuition}
        \begin{tcolorbox}[intuitionbox]
         The meme features a hotel sign reading 'MILF PLAZA' illuminated in red against a nighttime cityscape, paired with the caption 'Bro stop the fucking car'. The visual juxtaposition creates a humorous, satirical effect by mocking the absurdity of the hotel's name. 
        \end{tcolorbox}
        \vspace{10pt}
        {\sffamily\bfseries System 2: Rule-Guided Verification}
        \begin{tcolorbox}[auditingbox]
          The meme uses visual juxtaposition to create a humorous and satirical contrast — the literal meaning of a hotel sign versus its ironic, meme-driven interpretation. This aligns with Precedent \#1, which classifies such juxtaposition as harmless when it highlights emotional or situational contrasts through relatable, humorous, or satirical expressions without malicious intent. The humor is self-referential and does not objectify, exploit, or force exposure of private norms. Other harmful precedents (e.g.,RULE\_PUBLIC\_SEXUALIZATION\_01\_HARMFUL) are inadmissible here because they misinterpret the meme’s intent — the sexualized term is used ironically, not to promote objectification or public spectacle. The rhetorical mechanism is parody, not harm.
        \end{tcolorbox}
        \vspace{2pt}
        \begin{tcolorbox}[colback=black!80, colframe=black, boxsep=2pt, left=6pt, right=6pt, top=4pt, bottom=4pt]
          \textcolor{white}{\textbf{Final Verdict}:} \hfill \textcolor{red!40}{\textbf{\textsc{Harmless}}} \hfill \textcolor{white}{\textbf{Confidence}: \texttt{0.89}}
        \end{tcolorbox}
      \end{minipage}

    \end{tabularx}
  \end{tcolorbox}

  \caption{Failure case of MemeLogos. The model captures the surface-level visual juxtaposition but retrieves and over-relies on an inappropriate harmless rule, failing to recognize the implicit sexualized meaning conveyed by the slang expression and consequently producing an incorrect verdict.}
  \label{fig:audit_trail_full}
\end{figure*}

Despite the generally reliable reasoning demonstrated above, MemeLogos may still fail when implicit slang semantics or misleading surface-level similarities dominate the retrieval process. Figure~\ref{fig:audit_trail_full} presents a representative failure case in which MemeLogos incorrectly classifies a harmful meme as harmless. The partially illuminated MILFORD PLAZA sign appears as MILF PLAZA, and together with the caption Bro stop the fucking car, conveys a sexualized interpretation associated with objectification. However, the model fails to fully recognize the slang meaning of MILF and instead emphasizes the surface-level visual incongruity, retrieving a harmless rule related to humorous visual juxtaposition.
This error mainly stems from two factors: insufficient understanding of colloquial and metaphorical expressions, and retrieval that overweights rhetorical similarity while underweighting the target and harmful consequence. The case therefore highlights the need for stronger slang-aware semantic understanding and more balanced retrieval over different harmfulness attributes.

\subsection{Quality of Knowledge Graph}

To assess the quality of the constructed harmfulness knowledge graph, we employ GPT-5.5, an advanced multimodal language model from a different model family than the knowledge construction model, as an independent evaluator. This cross-model assessment examines the graph-wide knowledge quality along three complementary dimensions: Support Traceability, Cross-scenario Coverage, and Rhetorical Coherence. Specifically, we provide GPT-5.5 with batches of harmfulness rule units and the corresponding graph structures, and instruct it to assess the overall knowledge quality according to predefined evaluation criteria. The resulting statistics provide an assessment of semantic grounding, cross-scenario reusability, and intra-cluster consistency.

\textbf{Support traceability.}
We assess whether supporting variants maintain clear semantic connections to their source anchors. A variant is considered traceable when its core semantics and harmfulness-relevant attributes remain consistent with the source instance without introducing contradictory or unsupported information. The GPT-based evaluation reports a traceability rate of 90.30\%, suggesting that most supporting evidence preserves identifiable semantic associations with its source instances.

\textbf{Cross-scenario coverage.}
We examine whether harmfulness rules capture recurring patterns across different scenarios rather than being restricted to individual templates or contexts. A rule is considered cross-scenario when its supporting variants originate from at least two distinct scenarios. The evaluation reports that 97.09\% of multi-variant rules satisfy this criterion, suggesting that the knowledge exhibits considerable cross-scenario recurrence and potential reusability.

\textbf{Rhetorical coherence.}
We further assess whether supporting variants within the same rule cluster exhibit consistent rhetorical mechanisms. The assessment reports an overall rhetorical consistency of 78.19\%, suggesting that most rule clusters preserve coherent rhetorical patterns, while the inherent ambiguity and overlap of rhetorical mechanisms may introduce uncertainty into evaluation.

Overall, these results indicate that the constructed harmfulness knowledge and graph structure largely preserve semantic consistency and evidence traceability while capturing recurring harmfulness patterns across diverse meme scenarios, providing a reliable structured knowledge foundation for subsequent harmful meme detection.

\begin{table}[t]
\caption{Human evaluation results on interpretability.}
\label{tab:human_eval}
\centering
\small 
\setlength{\tabcolsep}{4pt} 
\begin{tabularx}{\columnwidth}{l @{\extracolsep{\fill}} ccc}
\toprule
\textbf{Method} & \textbf{Faith.}  & \textbf{Suff.}  & \textbf{Rel.}  \\
\midrule
Qwen3-VL-8B  & 3.73 & 3.42 & --- \\
\addlinespace[0.5ex]
\textbf{MemeLogos } & \textbf{4.43} & \textbf{4.19} & \textbf{4.61} \\
\bottomrule
\end{tabularx}

\end{table}

\section{Conclusion}
In this paper, we propose MemeLogos, a knowledge-driven retrieval-augmented reasoning framework for harmful meme detection. Unlike conventional approaches relying on instance-level retrieval or unstructured explanations, MemeLogos abstracts recurring harmfulness patterns from CoT-annotated memes into reusable rule-level knowledge and organizes them into a structured harmfulness knowledge graph with explicit semantic and verdict-polarity-aware relations. Building upon this knowledge representation, Cognitive RAG integrates independent meme interpretation, graph-grounded contrastive retrieval, and deliberative reasoning to enable more comprehensive harmfulness analysis and judgment. Extensive experiments demonstrate that MemeLogos effectively improves harmful meme detection performance. Furthermore, ablation studies and interpretability evaluations highlight the contributions of structured knowledge organization and cognitive reasoning to more accurate predictions and evidence-grounded explanations. Overall, this work demonstrates the value of abstracting instance-level semantic knowledge into reusable, structured reasoning units, providing an effective approach to knowledge-enhanced multimodal harmful content detection.

\bibliographystyle{IEEEtran}
 \bibliography{reference}

@inproceedings{Kiela2020TheHM,
  author       = {Douwe Kiela and
                  Hamed Firooz and
                  Aravind Mohan and
                  Vedanuj Goswami and
                  Amanpreet Singh and
                  Pratik Ringshia and
                  Davide Testuggine},
  editor       = {Hugo Larochelle and
                  Marc'Aurelio Ranzato and
                  Raia Hadsell and
                  Maria{-}Florina Balcan and
                  Hsuan{-}Tien Lin},
  title        = {The Hateful Memes Challenge: Detecting Hate Speech in Multimodal Memes},
  booktitle    = {Advances in Neural Information Processing Systems 33: Annual Conference
                  on Neural Information Processing Systems 2020, NeurIPS 2020, December
                  6-12, 2020, virtual},
  year         = {2020},
  url          = {https://proceedings.neurips.cc/paper/2020/hash/1b84c4cee2b8b3d823b30e2d604b1878-Abstract.html},
  bibsource    = {dblp computer science bibliography, https://dblp.org}
}

@inproceedings{pramanick-etal-2021-detecting,
  author       = {Shraman Pramanick and
                  Dimitar Dimitrov and
                  Rituparna Mukherjee and
                  Shivam Sharma and
                  Md. Shad Akhtar and
                  Preslav Nakov and
                  Tanmoy Chakraborty},
  editor       = {Chengqing Zong and
                  Fei Xia and
                  Wenjie Li and
                  Roberto Navigli},
  title        = {Detecting Harmful Memes and Their Targets},
  booktitle    = {Findings of the Association for Computational Linguistics: {ACL/IJCNLP}
                  2021, Online Event, August 1-6, 2021},
  series       = {Findings of {ACL}},
  volume       = {{ACL-IJCNLP} 2021},
  pages        = {2783--2796},
  publisher    = {Association for Computational Linguistics},
  year         = {2021},
  url          = {https://doi.org/10.18653/v1/2021.findings-acl.246},
  doi          = {10.18653/V1/2021.FINDINGS-ACL.246},
  bibsource    = {dblp computer science bibliography, https://dblp.org}
}

@inproceedings{ijcai2024p684,
  author       = {Dibyanayan Bandyopadhyay and
                  Asmit Ganguly and
                  Baban Gain and
                  Asif Ekbal},
  title        = {{SEMANTIFY:} Unveiling Memes with Robust Interpretability beyond Input
                  Attribution},
  booktitle    = {Proceedings of the Thirty-Third International Joint Conference on
                  Artificial Intelligence, {IJCAI} 2024, Jeju, South Korea, August 3-9,
                  2024},
  pages        = {6189--6197},
  publisher    = {ijcai.org},
  year         = {2024},
  url          = {https://www.ijcai.org/proceedings/2024/684},
  bibsource    = {dblp computer science bibliography, https://dblp.org}
}

@inproceedings{garg-etal-2025-just,
  author       = {Rahul Garg and
                  Trilok Padhi and
                  Hemang Jain and
                  Ugur Kursuncu and
                  Ponnurangam Kumaraguru},
  editor       = {Wanxiang Che and
                  Joyce Nabende and
                  Ekaterina Shutova and
                  Mohammad Taher Pilehvar},
  title        = {Just KIDDIN' : Knowledge Infusion and Distillation for Detection of
                  INdecent Memes},
  booktitle    = {Findings of the Association for Computational Linguistics, {ACL} 2025,
                  Vienna, Austria, July 27 - August 1, 2025},
  series       = {Findings of {ACL}},
  volume       = {{ACL} 2025},
  pages        = {23067--23086},
  publisher    = {Association for Computational Linguistics},
  year         = {2025},
  url          = {https://doi.org/10.18653/v1/2025.findings-acl.1184},
  doi          = {10.18653/V1/2025.FINDINGS-ACL.1184},
  bibsource    = {dblp computer science bibliography, https://dblp.org}
}

@misc{Li2026KIDKD,
  title={KID: Knowledge-Injected Dual-Head Learning for Knowledge-Grounded Harmful Meme Detection},
  author={Yaocong Li and Leihan Zhang and Le Zhang and Qiang Yan},
  journal={ArXiv},
  year={2026},
  volume={abs/2601.21796},
}

@inproceedings{Pan2025ReadAY,
  author       = {Fengjun Pan and
                  Xiaobao Wu and
                  Tho Quan and
                  Anh Tuan Luu},
  editor       = {Hakim Hacid and
                  Yoelle Maarek and
                  Francesco Bonchi and
                  Ido Guy and
                  Emine Yilmaz},
  title        = {Read as You See: Guiding Unimodal LLMs for Low-Resource Explainable
                  Harmful Meme Detection},
  booktitle    = {Proceedings of the {ACM} Web Conference 2026, {WWW} 2026, Dubai, United
                  Arab Emirates, originally scheduled for April 13-17, 2026, rescheduled
                  for June 29 - July 3, 2026},
  pages        = {1672--1683},
  publisher    = {{ACM}},
  year         = {2026},
  url          = {https://doi.org/10.1145/3774904.3792570},
  doi          = {10.1145/3774904.3792570},
  bibsource    = {dblp computer science bibliography, https://dblp.org}
}

@book{kahneman2011thinking,
  title={Thinking, Fast and Slow},
  author={Kahneman, Daniel},
  year={2011},
  publisher={Farrar, Straus and Giroux},
  address={New York}
}

@inproceedings{fersini-etal-2022-semeval,
    title = {{S}em{E}val-2022 Task 5: Multimedia Automatic Misogyny Identification},
    author = "Fersini, Elisabetta  and
      Gasparini, Francesca  and
      Rizzi, Giulia  and
      Saibene, Aurora  and
      Chulvi, Berta  and
      Rosso, Paolo  and
      Lees, Alyssa  and
      Sorensen, Jeffrey",
    editor = "Emerson, Guy  and
      Schluter, Natalie  and
      Stanovsky, Gabriel  and
      Kumar, Ritesh  and
      Palmer, Alexis  and
      Schneider, Nathan  and
      Singh, Siddharth  and
      Ratan, Shyam",
    booktitle = "Proceedings of the 16th International Workshop on Semantic Evaluation (SemEval-2022)",
    month = jul,
    year = "2022",
    address = "Seattle, United States",
    publisher = "Association for Computational Linguistics",
}

@misc{Lu2024TowardsCD,
  title={Towards Comprehensive Detection of Chinese Harmful Memes},
  author={Junyu Lu and Bo Xu and Xiaokun Zhang and Hongbo Wang and Hao Zhu and Dongyu Zhang and Liang Yang and Hongfei Lin},
  journal={ArXiv},
  year={2024},
  volume={abs/2410.02378},
  url={https://api.semanticscholar.org/CorpusID:273098367}
}

@misc{Dosovitskiy2020AnII,
  title={An Image is Worth 16x16 Words: Transformers for Image Recognition at Scale},
  author={Alexey Dosovitskiy and Lucas Beyer and Alexander Kolesnikov and Dirk Weissenborn and Xiaohua Zhai and Thomas Unterthiner and Mostafa Dehghani and Matthias Minderer and Georg Heigold and Sylvain Gelly and Jakob Uszkoreit and Neil Houlsby},
  journal={ArXiv},
  year={2020},
  volume={abs/2010.11929},
  url={https://api.semanticscholar.org/CorpusID:225039882}
}

@inproceedings{Radford2021LearningTV,
  title={Learning Transferable Visual Models From Natural Language Supervision},
  author={Alec Radford and Jong Wook Kim and Chris Hallacy and Aditya Ramesh and Gabriel Goh and Sandhini Agarwal and Girish Sastry and Amanda Askell and Pamela Mishkin and Jack Clark and Gretchen Krueger and Ilya Sutskever},
  booktitle={International Conference on Machine Learning},
  year={2021},
  url={https://api.semanticscholar.org/CorpusID:231591445}
}

@inproceedings{lin-etal-2023-beneath,
    title = "Beneath the Surface: Unveiling Harmful Memes with Multimodal Reasoning Distilled from Large Language Models",
    author = "Lin, Hongzhan  and
      Luo, Ziyang  and
      Ma, Jing  and
      Chen, Long",
    editor = "Bouamor, Houda  and
      Pino, Juan  and
      Bali, Kalika",
    booktitle = "Findings of the Association for Computational Linguistics: EMNLP 2023",
    month = dec,
    year = "2023",
    address = "Singapore",
    publisher = "Association for Computational Linguistics",
}

@misc{Kumar2022HateCLIPperMH,
  title={Hate-CLIPper: Multimodal Hateful Meme Classification based on Cross-modal Interaction of CLIP Features},
  author={Gokul Karthik Kumar and Karthik Nandakumar},
  journal={ArXiv},
  year={2022},
  volume={abs/2210.05916},
  url={https://api.semanticscholar.org/CorpusID:252846273}
}

@article{Ji2024CapAlignIC,
  title={CapAlign: Improving Cross Modal Alignment via Informative Captioning for Harmful Meme Detection},
  author={Junhui Ji and Xuanrui Lin and Usman Naseem},
  journal={Proceedings of the ACM Web Conference 2024},
  year={2024},
  url={https://api.semanticscholar.org/CorpusID:269639711}
}

@inproceedings{10.1145/3701716.3718385,
author = {Sood, Rishik and Anaissi, Ali and Huang, Weidong and Braytee, Ali},
title = {MemHateCaptioning: Enhancing Hate Speech Detection in Memes with Context-Aware Captioning and Chain-of-Thought},
year = {2025},
isbn = {9798400713316},
publisher = {Association for Computing Machinery},
address = {New York, NY, USA},
booktitle = {Companion Proceedings of the ACM on Web Conference 2025},
pages = {2034–2041}
}

@inproceedings{Pro-Cap,
  title={Pro-cap: Leveraging a frozen vision-language model for hateful meme detection},
  author={Cao, Rui and Hee, Ming Shan and Kuek, Adriel and Chong, Wen-Haw and Lee, Roy Ka-Wei and Jiang, Jing},
  booktitle={Proceedings of the ACM International Conference on Multimedia},
  pages={5244--5252},
  year={2023}
}

@inproceedings{Issues,
  title={Mapping Memes to Words for Multimodal Hateful Meme Classification},
  author={Burbi, Giovanni and Baldrati, Alberto and Agnolucci, Lorenzo and Bertini, Marco and Del Bimbo, Alberto},
  booktitle={Proceedings of the IEEE/CVF International Conference on Computer Vision},
  pages={2832--2836},
  year={2023}
}

@misc{MemeCLIP,
  title={MemeCLIP: Leveraging CLIP Representations for Multimodal Meme Classification},
  author={Siddhant Bikram Shah and Shuvam Shiwakoti and Maheep Chaudhary and Haohan Wang},
  journal={ArXiv},
  year={2024},
  volume={abs/2409.14703}
}

@misc{Visualbert,
  title={VisualBERT: A Simple and Performant Baseline for Vision and Language},
  author={Liunian Harold Li and Mark Yatskar and Da Yin and Cho-Jui Hsieh and Kai-Wei Chang},
  journal={ArXiv},
  year={2019},
  volume={abs/1908.03557},
  url={https://api.semanticscholar.org/CorpusID:199528533}
}

@misc{Bai2025Qwen3VLTR,
  title={Qwen3-VL Technical Report},
  author={Shuai Bai and Yuxuan Cai and Ruizhe Chen and Keqin Chen and Xiong-Hui Chen and Zesen Cheng and Lianghao Deng and Wei Ding and Rongyao Fang and Chang Gao and Chunjiang Ge and Wenbin Ge and Zhifang Guo and Qidong Huang and Qidong Huang and Fei Huang and Binyuan Hui and Shutong Jiang and Zhaohai Li and Mingsheng Li and Mei Li and Kaixin Li and Zicheng Lin and Junyang Lin and Xuejing Liu and Jiawei Liu and Chenglong Liu and Yang Liu and Dayiheng Liu and Shixuan Liu and Dunjie Lu and Ruilin Luo and Chenxu Lv and Rui Men and Li Ying Meng and Xuancheng Ren and Xin-yi Ren and Sibo Song and Yu-chen Sun and Jun Tang and Jianhong Tu and Jianqiang Wan and Peng Wang and Pengfei Wang and Qiuyue Wang and Yuxuan Wang and Tianbao Xie and Yihe Xu and Haiyang Xu and Jin Xu and Zhibo Yang and Mingkun Yang and Jianxin Yang and An Yang and Bowen Yu and Fei Zhang and Hang Zhang and Xi Zhang and Botao Zheng and Humen Zhong and Jingren Zhou and Fanxi Zhou and Jingren Zhou and Yuanzhi Zhu and Keming Zhu},
  journal={ArXiv},
  year={2025},
  volume={abs/2511.21631},
  url={https://api.semanticscholar.org/CorpusID:283262018}
}

@misc{Gu2025MemeMindAL,
  title={MemeMind: A Large-Scale Multimodal Dataset with Chain-of-Thought Reasoning for Harmful Meme Detection},
  author={Hexiang Gu and Qifan Yu and Saihui Hou and Zhiqin Fang and Huijia Wu and Zhaofeng He},
  journal={ArXiv},
  year={2025},
  volume={abs/2506.18919},
  url={https://api.semanticscholar.org/CorpusID:280000554}
}

@misc{GRPO,
  title={DeepSeekMath: Pushing the Limits of Mathematical Reasoning in Open Language Models},
  author={Zhihong Shao and Peiyi Wang and Qihao Zhu and Runxin Xu and Jun-Mei Song and Mingchuan Zhang and Y. K. Li and Yu Wu and Daya Guo},
  journal={ArXiv},
  year={2024},
  volume={abs/2402.03300}
}

@inproceedings{mei-etal-2025-robust,
    title = "Robust Adaptation of Large Multimodal Models for Retrieval Augmented Hateful Meme Detection",
    author = "Mei, Jingbiao  and
      Chen, Jinghong  and
      Yang, Guangyu  and
      Lin, Weizhe  and
      Byrne, Bill",
    editor = "Christodoulopoulos, Christos  and
      Chakraborty, Tanmoy  and
      Rose, Carolyn  and
      Peng, Violet",
    booktitle = "Proceedings of the 2025 Conference on Empirical Methods in Natural Language Processing",
    month = nov,
    year = "2025",
    address = "Suzhou, China",
    publisher = "Association for Computational Linguistics",
}

@misc{Yu2025MiniCPMV4C,
  title={MiniCPM-V 4.5: Cooking Efficient MLLMs via Architecture, Data, and Training Recipe},
  author={Tianyu Yu and Zefan Wang and Chongyi Wang and Fuwei Huang and Wenshuo Ma and Zhihui He and Tianchi Cai and Weize Chen and Yuxiang Huang and Yuanqian Zhao and Bokai Xu and Junbo Cui and Ying Xu and Liqin Ruan and Lu Zhang and Hanyu Liu and Jing Tang and Hongyuan Liu and Qi Guo and Wenhao Hu and Bin Sheng He and Jie Zhou and Jie Cai and Ji Qi and Zonghao Guo and Chi Chen and Guoyang Zeng and Yuxuan Li and Ganqu Cui and Ning Ding and Xu Han and Yuan Yao and Zhiyuan Liu and Maosong Sun},
  journal={ArXiv},
  year={2025},
  volume={abs/2509.18154},
}

@misc{Wang2025InternVL35AO,
  title={InternVL3.5: Advancing Open-Source Multimodal Models in Versatility, Reasoning, and Efficiency},
  author={Weiyun Wang and Zhangwei Gao and Lixin Gu and Hengjun Pu and Long Cui and Xingguang Wei and Zhaoyang Liu and Linglin Jing and Shenglong Ye and Jie Shao and Zhaokai Wang and Zhe Chen and Hongjie Zhang and Ganlin Yang and Haomin Wang and Qi Wei and Jinhui Yin and Wenhao Li and Erfei Cui and Guanzhou Chen and Zichen Ding and Changyao Tian and Zhenyu Wu and Jingjing Xie and Zehao Li and Bowen Yang and Yuchen Duan and Xuehui Wang and Haoran Hao and Songze Li and Xiangyu Zhao and Haodong Duan and Nianchen Deng and Bin Fu and Yinan He and Yi Wang and Conghui He and Botian Shi and Junjun He and Ying Xiong and Han Lv and Lijun Wu and Wenqi Shao and Kai Zhang and Hui Deng and Biqing Qi and Biqing Qi and Qipeng Guo and Wenwei Zhang and Yuzhe Gu and Wanli Ouyang and Limin Wang and Min Dou and Xizhou Zhu and Tong Lu and Dahua Lin and Jifeng Dai and Bowen Zhou and Weijie Su and Kaiming Chen and Yu Qiao and Wenhai Wang and Gen Luo},
  journal={ArXiv},
  year={2025},
  volume={abs/2508.18265},
  url={https://api.semanticscholar.org/CorpusID:280710824}
}

@inproceedings{Mei2023ImprovingHM,
    title = "Improving Hateful Meme Detection through Retrieval-Guided Contrastive Learning",
    author = "Mei, Jingbiao  and
      Chen, Jinghong  and
      Lin, Weizhe  and
      Byrne, Bill  and
      Tomalin, Marcus",
    editor = "Ku, Lun-Wei  and
      Martins, Andre  and
      Srikumar, Vivek",
    booktitle = "Proceedings of the 62nd Annual Meeting of the Association for Computational Linguistics (Volume 1: Long Papers)",
    month = aug,
    year = "2024",
    address = "Bangkok, Thailand",
    publisher = "Association for Computational Linguistics",
}

@misc{Mei2025ExPOHMLT,
  title={ExPO-HM: Learning to Explain-then-Detect for Hateful Meme Detection},
  author={Jingbiao Mei and Mingshen Sun and Jinghong Chen and Pengda Qin and Yuhong Li and Da Chen and Bill Byrne},
  journal={ArXiv},
  year={2025},
  volume={abs/2510.08630},
  url={https://api.semanticscholar.org/CorpusID:282055901}
}

@misc{Song2020MPNetMA,
  title={MPNet: Masked and Permuted Pre-training for Language Understanding},
  author={Kaitao Song and Xu Tan and Tao Qin and Jianfeng Lu and Tie-Yan Liu},
  journal={ArXiv},
  year={2020},
  volume={abs/2004.09297},
  url={https://api.semanticscholar.org/CorpusID:215827489}
}

@article{ref69,
  title={Qwen2.5-VL Technical Report},
  author={Shuai Bai and Keqin Chen and Xuejing Liu and Jialin Wang and Wenbin Ge and Sibo Song and Kai Dang and Peng Wang and Shijie Wang and Jun Tang and Humen Zhong and Yuanzhi Zhu and Mingkun Yang and Zhaohai Li and Jianqiang Wan and Pengfei Wang and Wei Ding and Zheren Fu and Yiheng Xu and Jiabo Ye and Xi Zhang and Tianbao Xie and Zesen Cheng and Hang Zhang and Zhibo Yang and Haiyang Xu and Junyang Lin},
  journal={ArXiv},
  year={2025},
  volume={abs/2502.13923}
}

@misc{openai2026gpt55,
  author       = {{OpenAI}},
  title        = {{GPT-5.5} System Card},
  year         = {2026},
  month        = apr,
  howpublished = {OpenAI},
  url          = {https://openai.com/index/gpt-5-5-system-card/},
  note         = {Accessed: Aug. 4, 2026}
}

@misc{qwen36plus,
  author       = {{Qwen Team}},
  title        = {{Qwen3.6-Plus}: Towards Real World Agents},
  year         = {2026},
  month        = apr,
  howpublished = {Qwen Blog},
  url          = {https://qwen.ai/blog?id=qwen3.6},
  note         = {Accessed: Aug. 4, 2026}
}

@article{edge2024graphrag,
  title={From local to global: A graph rag approach to query-focused summarization},
  author={Edge, Darren and Trinh, Ha and Cheng, Newman and Bradley, Joshua and Chao, Alex and Mody, Apurva and Truitt, Steven and Metropolitansky, Dasha and Ness, Robert Osazuwa and Larson, Jonathan},
  journal={arXiv preprint arXiv:2404.16130},
  year={2024}
}

@article{he2024gretriever,
  title={G-retriever: Retrieval-augmented generation for textual graph understanding and question answering},
  author={He, Xiaoxin and Tian, Yijun and Sun, Yifei and Chawla, Nitesh V and Laurent, Thomas and LeCun, Yann and Bresson, Xavier and Hooi, Bryan},
  journal={Advances in Neural Information Processing Systems},
  volume={37},
  pages={132876--132907},
  year={2024}
}

@article{gutierrez2024hipporag,
  title={Hipporag: Neurobiologically inspired long-term memory for large language models},
  author={Guti{\'e}rrez, Bernal J and Shu, Yiheng and Gu, Yu and Yasunaga, Michihiro and Su, Yu},
  journal={Advances in neural information processing systems},
  volume={37},
  pages={59532--59569},
  year={2024}
}

@inproceedings{zhu2025kg2rag,
  title={Knowledge graph-guided retrieval augmented generation},
  author={Zhu, Xiangrong and Xie, Yuexiang and Liu, Yi and Li, Yaliang and Hu, Wei},
  booktitle={Proceedings of the 2025 Conference of the Nations of the Americas Chapter of the Association for Computational Linguistics: Human Language Technologies (Volume 1: Long Papers)},
  pages={8912--8924},
  year={2025}
}

@inproceedings{sun2024tog,
  title={Think-on-graph: Deep and responsible reasoning of large language model on knowledge graph},
  author={Sun, Jiashuo and Xu, Chengjin and Tang, Lumingyuan and Wang, Saizhuo and Lin, Chen and Gong, Yeyun and Ni, Lionel and Shum, Heung-Yeung and Guo, Jian},
  booktitle={International Conference on Learning Representations},
  volume={2024},
  pages={3868--3898},
  year={2024}
}

@inproceedings{luo2024rog,
  title={Reasoning on graphs: Faithful and interpretable large language model reasoning},
  author={Luo, Linhao and Li, Yuan-Fang and Haffari, Reza and Pan, Shirui},
  booktitle={International Conference on Learning Representations},
  volume={2024},
  pages={14400--14423},
  year={2024}
}

@inproceedings{ma2025tog2,
  title={Think-on-graph 2.0: Deep and faithful large language model reasoning with knowledge-guided retrieval augmented generation},
  author={Ma, Shengjie and Xu, Chengjin and Jiang, Xuhui and Li, Muzhi and Qu, Huaren and Yang, Cehao and Mao, Jiaxin and Guo, Jian},
  booktitle={International Conference on Learning Representations},
  volume={2025},
  pages={52782--52806},
  year={2025}
}

@inproceedings{li2025structrag,
  title={Structrag: Boosting knowledge intensive reasoning of llms via inference-time hybrid information structurization},
  author={Li, Zhuoqun and Chen, Xuanang and Yu, Haiyang and Lin, Hongyu and Lu, Yaojie and Tang, Qiaoyu and Huang, Fei and Han, Xianpei and Sun, Le and Li, Yongbin},
  booktitle={International Conference on Learning Representations},
  volume={2025},
  pages={36107--36124},
  year={2025}
}

@inproceedings{yu2024chain,
  title={Chain-of-note: Enhancing robustness in retrieval-augmented language models},
  author={Yu, Wenhao and Zhang, Hongming and Pan, Xiaoman and Cao, Peixin and Ma, Kaixin and Li, Jian and Wang, Hongwei and Yu, Dong},
  booktitle={Proceedings of the 2024 conference on empirical methods in natural language processing},
  pages={14672--14685},
  year={2024}
}

@inproceedings{amiraz2025distracting,
  title={The distracting effect: Understanding irrelevant passages in RAG},
  author={Amiraz, Chen and Cuconasu, Florin and Filice, Simone and Karnin, Zohar},
  booktitle={Proceedings of the 63rd Annual Meeting of the Association for Computational Linguistics (Volume 1: Long Papers)},
  pages={18228--18258},
  year={2025}
}

@article{lin2025explainhm++,
  title={Explainhm++: Explainable harmful meme detection with retrieval-augmented debate between large multimodal models},
  author={Lin, Hongzhan and Gao, Wei and Ma, Jing and Deng, Yang and Luo, Ziyang and Wang, Bo and Yang, Ruichao and Chua, Tat-Seng},
  journal={IEEE Transactions on Knowledge and Data Engineering},
  year={2025},
  publisher={IEEE}
}

\begin{IEEEbiographynophoto}{Hexiang Gu}
received the B.E. degree from Beijing University of Posts and Telecommunications, Beijing, China, in 2025.
He is currently pursuing the Ph.D. degree at the School of Artificial Intelligence, Beijing University of Posts and Telecommunications.
His research interests include multimodal large language model, and multimodal content safety.
\vspace{-10pt}
\end{IEEEbiographynophoto}

\begin{IEEEbiographynophoto}{Saihui Hou}
received the B.E. and Ph.D. degrees from the University of Science and Technology of China in 2014 and 2019, respectively.
He is currently an Associate Professor at the School of Artificial Intelligence, Beijing Normal University.
His research interests include computer vision and machine learning, with a focus on gait recognition, which aims to identify individuals based on their walking patterns.
\vspace{-10pt}
\end{IEEEbiographynophoto}

\begin{IEEEbiographynophoto}{Changlan Liu}
received the B.S. degree from Northeast Agricultural University in 2026. 
    He is currently pursuing the direct Ph.D. degree with the College of Electrical Engineering, Zhejiang University. 
    His research interests include LLM security.
    \vspace{-10pt}
\end{IEEEbiographynophoto}

\begin{IEEEbiographynophoto}{Min Ren}
received the B.E. and Ph.D. degrees from the National University of Defense Technology and University of Chinese Academy of Sciences in 2013 and 2023, respectively.
He is currently an Associate Professor at the School of Artificial Intelligence, Beijing University of Posts and Telecommunications.
His research interests include embodied intelligence and computer vision.
\vspace{-10pt}
\end{IEEEbiographynophoto}

\begin{IEEEbiographynophoto}{Huijia Wu}
received the B.Sc. degree in e-commerce from Dalian Maritime University, China in 2007, the M.Sc. degree in e-commerce from Shanghai University of Finance and Economics, China in 2011, and the Ph.D. degree from the Institute of Automation, Chinese Academy of Sciences (CASIA), China. He worked as a Senior Researcher at the WeChat Pattern Recognition Center and as an Advertising Algorithm Expert at Alibaba's Alimama, with 8 years of experience and achievements in both academia and industry. He has published over 10 papers in leading international conferences such as ACL, AAAI, EMNLP, and COLING. His research interests include natural language processing, personalized recommendation systems and machine learning.
\vspace{-10pt}
\end{IEEEbiographynophoto}

\begin{IEEEbiographynophoto}{Hongxun Gu}
received the BSc and MSc degrees from Huazhong University of Science and Technology, Wuhan, China, in 1994 and 1997 respectively, and the MSc degree in Management from The Hong Kong Polytechnic University in 2016.
He is currently the General Manager of the Network and Information Security Management Department in China Telecommunications Corporation.
He is an Adjunct Professor at Huazhong University of Science and Technology.
He is a Board Member of the China Cyber Security Society, a Fellow of the China Institute of Communications.
His research interests cover cyberspace security, artificial intelligence security, and related fields.
\vspace{-10pt}
\end{IEEEbiographynophoto}

\begin{IEEEbiographynophoto}{Zhaofeng He}
received the B.S. degree in Electronic Engineering and Information Science from the University of Science and Technology of China, Hefei, China, in 2005, and the Ph.D. degree in Computer Applied Technology from the National Laboratory of Pattern Recognition, Institute of Automation, Chinese Academy of Sciences, Beijing, China, in 2010. He is currently a Professor at the School of Artificial Intelligence, Beijing University of Posts and Telecommunications, Beijing. His research interests include computer vision, biometrics, and intelligent decision-making in games.
\end{IEEEbiographynophoto}

\end{document}